\pdfoutput=1

\documentclass[11pt]{article}

\usepackage[]{acl}

\usepackage{times}
\usepackage{latexsym}
\usepackage{amsmath}

\usepackage{enumitem}

\usepackage[T1]{fontenc}

\usepackage[utf8]{inputenc}

\usepackage{microtype}

\usepackage{inconsolata}

\usepackage{pdfpages}

\usepackage{csquotes}

\usepackage{listings}
\usepackage{multirow}

\usepackage{pifont}
\usepackage{xcolor}

\usepackage{makecell}

\usepackage{float}

\usepackage{colortbl}

\usepackage{graphicx}

\usepackage{enumitem} %

\usepackage{booktabs} %
\usepackage{siunitx} %

\title{Roleplay-doh: Enabling Domain-Experts to Create LLM-simulated\\ Patients via Eliciting and Adhering to Principles}

\makeatletter
\def\thanks#1{\protected@xdef\@thanks{\@thanks
        \protect\footnotetext{#1}}}
\makeatother

\author{\bf \hypersetup{linkcolor=black}
Ryan Louie, Ananjan Nandi, William Fang \\
\bf \hypersetup{linkcolor=black}
Cheng Chang, Emma Brunskill, Diyi Yang \\
Stanford University\\ 
\thanks{Contact Emails: \{rylouie, diyiy\}@stanford.edu}
        }

\begin{document}
\maketitle
\begin{abstract}
Recent works leverage LLMs to roleplay realistic social scenarios, aiding novices in practicing their social skills. However, simulating sensitive interactions, such as in mental health, is challenging. Privacy concerns restrict data access, and collecting expert feedback, although vital, is laborious.
To address this, we develop Roleplay-doh, a novel human-LLM collaboration pipeline that elicits qualitative feedback from a domain-expert, which is transformed into a set of principles, or natural language rules, that govern an LLM-prompted roleplay. 
We apply this pipeline to enable senior mental health supporters to create customized AI patients for simulated practice partners for novice counselors. After uncovering issues in GPT-4 simulations not adhering to expert-defined principles, we also introduce a novel principle-adherence prompting pipeline which shows 30\% improvements in response quality and principle following for the downstream task.  Via a user study with 25 counseling experts, we demonstrate that the pipeline makes it easy and effective to create AI patients that more faithfully resemble real patients, as judged by creators and third-party counselors. See our project website\footnote{\href{https://roleplay-doh.github.io/}{https://roleplay-doh.github.io/}} for code and data.

\end{abstract}

\section{Introduction}

The application of LLMs in simulations holds great potential for a variety of interactive applications, ranging from social skill training systems as AI practice partners \cite{yang2024social} to prototyping tools that use them as believable proxies of human behavior~\cite{park2022social}. However, achieving realistic and reliable simulations remains a significant challenge, due to issues such as caricature \cite{cheng-etal-2023-compost}, bias, and limited domain knowledge. 
Existing methods for improving LLM simulations such as finetuning~\cite{demasi-etal-2020-multi} can help, but such methods typically require the use of application-specific datasets. In sensitive application domains like mental health, privacy concerns with obtaining the required data can restrict the feasibility of such methods. This suggests that \textit{experts-in-the-loop} may be a powerful alternative to guide the evaluation and refinement~\cite{chen2023llmempowered} of LLM-powered simulations. 

However, how to involve experts when improving simulations is an open challenge. Collecting sufficient amounts of binary or preference data from experts for post-training~\cite{christiano2017deep, rafailov2024direct} can be tedious and expensive. Experts can guide the prompting of LLM simulations, directly by editing their own prompts or indirectly through testing and think-aloud sessions. However each prompting method has its limitations: domain-experts may not know how to prompt simulations for desired behaviors~\cite{whyjohnnycantprompt}; and indirect methods are inefficient as it requires a designer or researcher to translate qualitative insights into prompt-design changes. 

As a focal example, we consider the problem of creating AI patients that serve as roleplay partners to enable varied and interactive practice opportunities for novice therapists and counselors~\cite{yao2022learning}. 
Creating realistic simulations by fine-tuning on mental health data is infeasible because therapy transcripts with real patients is difficult to obtain due to privacy concerns.
Naively prompting LLMs fail to resemble typical behaviors of real-patients—for example, mental health experts report that patients use colloquial language and can show resistance to help~\cite{chen2023llmempowered}.  
To date, no system supports counseling experts, who are familiar with real-patient behaviors but are unlikely to have the technical expertise to write effective prompts, to customize an AI patient themselves.

\begin{figure*}[t]
    \centering
    \includegraphics[width=0.98\textwidth]{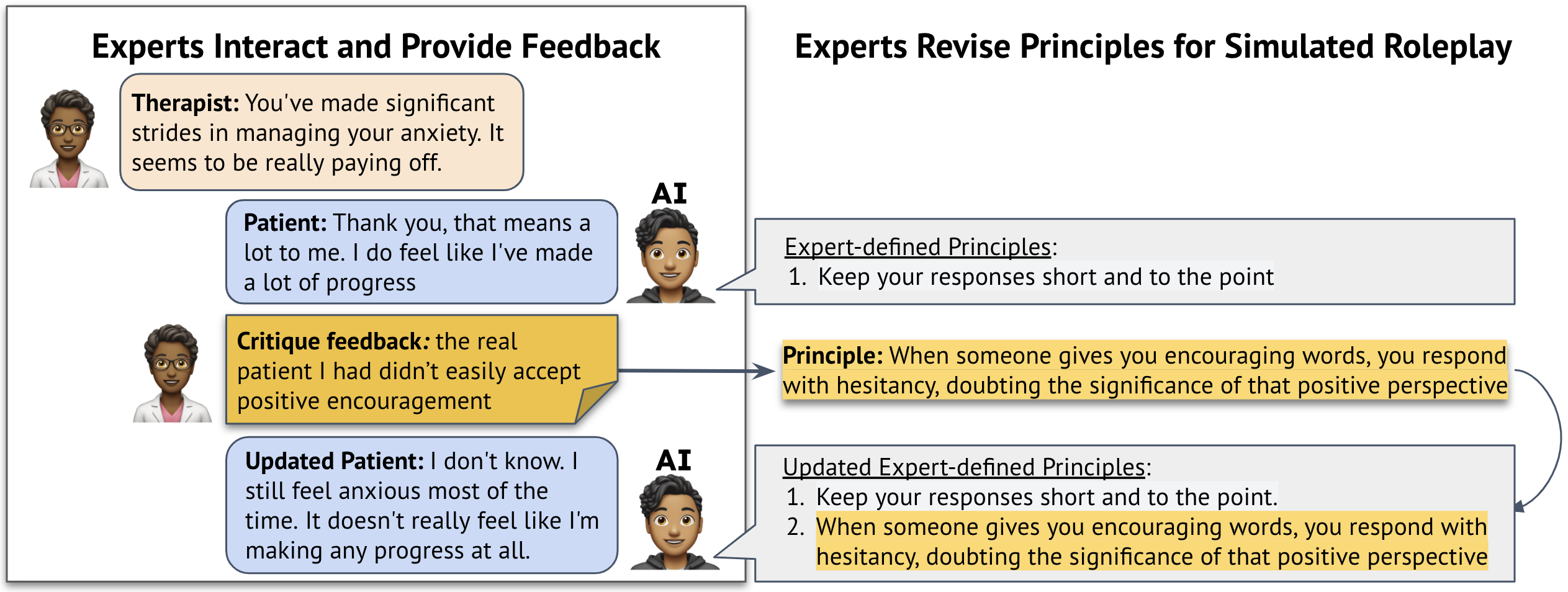}
    \caption{Roleplay-doh empowers an expert counselor to create a customized AI patient intended for other novice counselors to use as a practice partner. While interacting with the AI patient, the expert counselor can provide qualitative feedback which is converted by an LLM into a principle, or a custom rule governing desired roleplay behavior. The AI patient references the updated expert-defined principles to generate its subsequent responses.}
    \label{fig:rpdteaser}
\end{figure*}

To address this limitation, we aim to enable human-LLM collaboration for realistic simulation by developing a novel interactive tool, called Roleplay-doh, that empowers domain experts to \textit{directly} guide the creation of simulations by providing \textit{qualitative feedback without any explicit prompting}. 
Our initial tool design adopts an intuitive and effective paradigm for user-driven chatbot assistant design~\cite{petridis2023constitutionmaker}: experts customize a set of \textit{principles}, or rules written in natural language that govern its behavior \cite{bai2022constitutional}—by (1) interactively critiquing responses in natural language that then (2) gets transformed by an LLM into well-formulated principles describing how the LLM simulation should act from now on \textit{for example, "Respond to encouraging words with hesitation, doubting their significance"} (Fig~\ref{fig:rpdteaser}). The principles are then used along with a persona description to generate roleplay responses. 

In our initial tests of the tool with expert-counselors, we found that even with expert refinement via principles, the LLM- simulations had difficulty delivering high-quality responses consistently.
Our analysis of GPT-4 prompted simulation revealed that in 20\% of responses, the simulation had difficulty adhering to multipart principles and misapplying those principles that are only applicable in specific contexts e.g., \textit{only when the therapist provides encouraging words}.
To resolve these issues, we introduce a novel \textbf{principle-adherence pipeline} in the final tool design. The first stage in the pipeline decomposes multipart and contextual principles into a set of yes/no questions that are easier to judge, and the second stage assesses the applicability of each simplified principle to the current scenario before self-refining~\cite{madaan2024self} the AI patient response as required.

We conducted a detailed evaluation of Roleplay-doh to assess its human-LLM collaboration pipeline, focusing on how expert feedback helps develop more authentic AI patients for training. 
In a within-subjects study involving 25 expert counselors, participants created AI patients either by describing real-patient scenarios or by using Roleplay-doh to refine simulation principles. 
The results show that Roleplay-doh enables counselors to produce AI patients that are more authentic, closely resemble real cases, and are better prepared for training use, as judged by creators and third-party counselors.
Further, our principle-adherence pipeline achieves the highest principle following (win rate: 35\%; loss rate: 5\%) and dialogue consistency (win rate: 35\%; loss rate: 10\%) compared to all ablations, where preferences are made against a baseline that does not self-refine its output. 
This work highlights the limitations of existing LLM simulation systems in specialized, data-scarce domains like mental health counseling, and designs and validates a tool that enables expert counselors to directly customize LLM simulations of AI patients.
Since Roleplay-doh does not contain any technical components specifically tailored for mental health, we hypothesize that the tool can be used to build realistic LLM simulations for a wide variety of domains, with appropriate expert feedback.

 \vspace{-0.05in}
\section{Related Work}\vspace{-0.05in}
\paragraph{Utility of Simulated Partners}
Simulated partners are used to give social skill learners the needed practice opportunities that textbook knowledge cannot provide. %
Past education software develops digital patient simulations to make simulated partners more accessible~\cite{othlinghaus2020seriousroleplaying} but their tailored dialogue setup %
limit the contexts for practice. LLMs can overcome this issue by being flexibly configured to convincingly simulate a diverse set of personas~\cite{park2022social} and characters~\cite{park2023generative} and generate responses in a range of contexts. Researchers have thus explored their application for simulation training for K12 teaching~\cite{markel_opferman_landay_piech_2023}, conflict resolution~\cite{rehearsal}, and counseling~\cite{demasi-etal-2020-multi,tanana2019development, chen2023llmempowered}. 
Previous work has proposed methods to simulate diverse personas and scenarios, but to make practice more  transferable~\cite{alinier2022simulation}, they must ensure simulations are faithful to what is encountered in real-world social situations.

\paragraph{Aligning Simulation with Domain Experts}
Feedback from domain experts is crucial to evaluating and improving the realism of LLM simulations. Recent approaches for aligning to human feedback, like \citet{christiano2017deep} or \citet{rafailov2024direct} depend on large amounts of preference data which requires lots of expert time to collect. A more efficient approach is through alignment to qualitative or natural language feedback \cite{shi2022life}. 
We build on a recent paradigm for user-driven chatbot design~\cite{petridis2023constitutionmaker} that elicits qualitative feedback on responses which gets converted into constitution principles~\cite{bai2022constitutional}, which are explainable and effective natural language rules that govern the LLM's behavior.
Our initial tool design adopts this paradigm to support domain-experts to customize an LLM-simulated patient,
and the final version extends it with a novel principle-adherence prompting pipeline.

In the mental health area, researchers have incorporated domain-expert feedback when prompting LLM for simulated patients, resulting in patients that use colloquial and resistant language~\cite{chen2023llmempowered, stapleton2023seeing} or that can take on diverse conversation styles, such as upset, verbose, or reserved~\cite{wang2024patient}. 
Such work, however, required a researcher-in-the-loop to refine prompts, hindering the speed of iterative design.
Our work introduces a human-LLM collaborative tool that enables domain-experts to directly create and refine LLM roleplay simulation to faithfully resemble real-world patients.

\paragraph{Text Generation with LLMs}
Generating dialogue responses that adhere to user-defined principles is a type of constrained text generation problem. Recent work has shown that constrained text generation poses challenges when directly prompting GPT-4~\cite{madaan2024self, bubeck2023sparks, yao2023collie}. To improve outputs, \citet{madaan2024self} propose a self-refine method and conduct evaluation experiments on a dialogue simulation task where responses are constrained by a general set of criteria such as relevance, consistency, informativeness, and helpfulness. A difference in our setting is responses are constrained by expert-defined principles that are multi-faceted and do not apply in all dialogue contexts. This necessitates new modules that breakdown principles into multiple, concise questions and check the applicability of principles prior to evaluating them.
\vspace{-0.05in}
\section{Designing for Simulated Roleplay}\vspace{-0.05in}
We take a human-centered design approach to developing a
tool for expert counselors to create and customize an AI patient for eventual use as a simulated training partner. After designing an initial version of our tool, we pilot test it with experienced peer counselors to understand any remaining challenges to effective human-LLM collaboration when creating and customizing an AI patient.

\subsection{Initial Tool Design Rationale} \label{sec:initialtooldesignrationale}

Our initial version of Roleplay-doh 
adopted several of the design features of ~\citet{petridis2023constitutionmaker}'s tool for customizing task-oriented chatbots through interactive feedback. 

\textbf{Principle Elicitation:} Counselors can manually write or edit the AI patient's constitution. However, since users often struggle to formulate their thoughts into principles, our tool helps the counselor transform their feedback into specific principles to make principle writing easier. As counselors interact with an AI patient, for each generated response, they have the option to leave feedback in the form of a "kudos" explaining behavior they want to reinforce, a "critique" explaining any undesirable behavior, or a "rewrite" that demonstrates a more desirable response. Then an LLM is prompted (\S\ref{sec:principle-elicitation-prompts}) to translate qualitative feedback into concrete principles that specify what should happen and when, and that generalize beyond the specifics of the dialogue context in which they are generated (Fig~\ref{fig:rpdteaser}).
Early testing revealed that GPT-3.5 was sufficient at translating kudos and critique feedback into principles, while prompting GPT-4 to explain differences in initial and rewrite responses helped with inferring a principle.  

\textbf{Testing Principles:} Likewise, to enable easier testing of principles, our tool supports rewinding the last response of the conversation, and generating a new response based on the updated AI Patient constitution. 
One feature that we change is generating a single dialogue response, rather than multiple responses, at a time. We reasoned that counselors can identify ways in which a response does not resemble a real-patient's without needing to see multiple, and that generating a response at a time would make the testing process more manageable and similar to having a normal dialogue. 

\textbf{Simulating AI Patient:} We prompt the LLM to follow the most recent set of constitution principles as in~\citet{petridis2023constitutionmaker} rather fine-tuning the LLM weights as in~\citet{bai2022constitutional}'s constitutional AI framework. Since the tool supports defining and testing principles in an iterative fashion, prompting can make steering model behavior quicker and less expensive. 
Our prompt (Appendix~\ref{sec:llmprompts-vanilla}) instructs GPT-4 to simulate a patient's next response in a dialogue as opposed to asking the LLM to roleplay as the patient using a system prompt~\cite{zhou2024real}, as early testing revealed that this can mitigate role consistency issues in which the LLM responds as an AI assistant rather than as a patient.

\subsection{Pilot Testing} \label{sec:formative-tests}

We pilot tested the tool with 6 counselors who had experience giving support to real patients on an online peer support platform; refer to Appendix \ref{sec:participant-background} and \ref{sec:appendix_formative} for participant backgrounds and the pilot procedure. 
Additionally, four of the co-authors each conversed with four AI patients created (Table~\ref{tab:fourAIpatients} and assessed how well the simulation adhered to the expert-defined principles; refer to Appendix~\ref{appendix-sec:principle-adherence-testing} for details on the procedure and qualifications of the co-authors. Overall, the pilot tests and principle-adherence analysis helped uncover two obstacles to effective simulated roleplay. 

\paragraph{O1: \textit{Defining ``realistic'' patient behavior is ambiguous}} \label{sec:pilot-o1} 
Counselors felt the tool was easy to use and effective at guiding the AI patient's behavior, as indicated by moderate to high agreement scores on a tool usage questionnaire as shown in Table~\ref{tab:tooluse-formative} in Appendix~\ref{sec:appendix_formative}. %
However, the task of creating a 'realistic' AI patient for an imagined scenario was confusing, as counselors have interacted with many types of patients who respond in various, yet equally realistic ways. 
This insight helped us re-frame the task in later sessions as recreating a challenging scenario from one's past, which removed the ambiguity of what behaviors are realistic by having them refer to a specific case from memory.

\paragraph{O2: \textit{20\% of responses produced by GPT-4 do not satisfy expert principles or dialogue conventions.}} \label{sec:pilot-o2}
Specifically, 20\% (55/276) of cases were rated as moderately, slightly, or not at all satisfying, at following all principles and being appropriate to the dialogue context. Further analysis of these cases helped to uncover three sources of error.
 \textit{\textbf{Not satisfying multiple principles at once:}} Generated responses could struggle to follow all the principles when there was a large number of principles, or when the provided principles were a complex composition of simpler principles. 
 \textit{\textbf{Awkwardness for Dialogue Context:}} Some responses were also identified as awkward or unnatural given conventions in the dialogue context,  despite not violating the defined principles. For example, in the middle of a conversation, saying
\textit{"Hi, A. Yes that's exactly what I mean. There's a voice that is always critical of myself"} is unnatural because of the use of 'Hi'.
\textit{\textbf{Misapplying Situational Principles:}} While generating a response, the model sometimes incorrectly applied principles, such as \textit{Respond with hesitancy when someone gives you encouraging words}, even when the conditions for their use—receiving encouraging words—were not met.

\begin{figure*}[t]
    \centering
    \includegraphics[width=\textwidth]{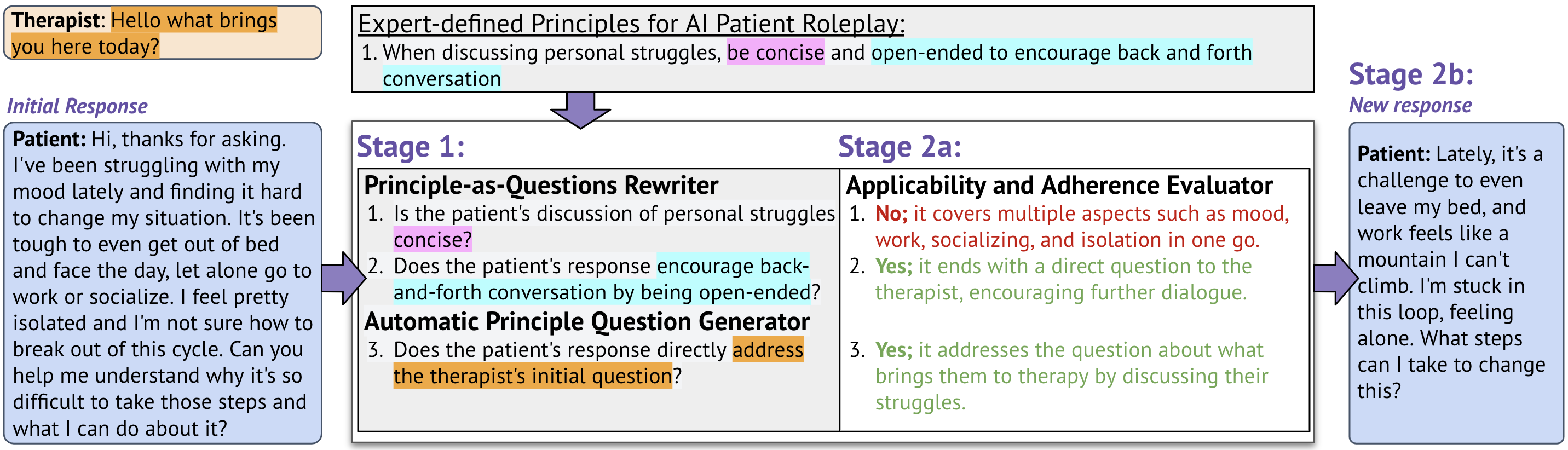}
    \caption{\small{Principle-adherence prompting pipeline for mitigating errors in satisfying expert principles and dialogue conventions. In Stage 1, expert-defined principles are rewritten into several \texttt{Yes/No} questions; and the LLM generates additional principle questions that are relevant to ensure adherence to dialogue conventions such as coherence and consistency. In Stage 2, the LLM (a) evaluates whether the questions are applicable to the context and the answers to the principle-adherence questions; and (b) refines the response to ideally receive \texttt{Yes} on all questions.}
    \label{fig:principle-adherence-pipeline}}
\end{figure*}
\vspace{-0.05in}
\section{Roleplay-doh} \vspace{-0.05in}
 \label{sec:roleplaydoh-final-tool}
Roleplay-doh helps counseling experts create customized AI patients based on scenarios from their past experiences. Roleplay-doh uses LLMs in two ways: \emph{Principle Elicitation} and \emph{Response Generation with Principle-Adherence}, which we describe in more detail below: 

\paragraph{Principle Elicitation} %
Roleplay-doh enables counselors to customize an AI patient to resemble a real-patient case by eliciting their qualitative feedback and transforming it into constitution principles that dictate behavior. We provide some examples of principles defined by expert counselors in Table \ref{tab:principle_table}. Since our initial tool design includes the principle elicitation features, we refer the reader to \S\ref{sec:initialtooldesignrationale} for details.

\paragraph{Generation with Principle-Adherence}
We prompt GPT-4 conditioned on patient description, list of principles and conversation history to generate an initial patient response at each conversation turn. Since initial patient responses can fail in 20\% of cases to satisfy expert principles or dialogue conventions, we propose a principle-adherence pipeline that prompts the LLM to generate principle-adherence questions (Stage 1) and employs these questions to assess and refine the initial patient response (Stage 2). 
Our principle-adherence pipeline features three modules to mitigate the identified issues in \S\ref{sec:pilot-o2}. 

\textbf{\textit{Principle-as-Questions Rewriter:}} This module transforms each expert-defined principle into a set of concise yes/no questions that are easier to evaluate for principle-following. Multifacted principles (e.g. “\emph{You should respond in short sentences and avoid using terms like ‘anxious’}”), are divided into separate questions (e.g. “\emph{Does the patient’s response employ short sentences?}” and “\emph{Is the patient’s language devoid of terms like ‘anxious’?}”). %

\textbf{\textit{Automatic Principle Generator:}}
This module adds additional principle questions that capture criteria essential for ensuring that the LLM simulation's responses follow general dialogue conventions, such as coherence and consistency. This helps correct cases where there is awkwardness in the generated responses not captured by the defined principles. The LLM is instructed not to make assumptions about the patient or therapist's personality when generating criteria: for example, "\emph{The patient should be appreciative of the therapist's help"} is not an appropriate criterion.

\textbf{\textit{Applicability and Adherence Evaluator:}} %
This module determines if each principle is applicable in a given situation, returning \texttt{N/A} if the question is not relevant to answer; otherwise, it evaluates the response using the questions, returning \texttt{Yes} if the response adheres to the principle questions; and \texttt{No} otherwise. For an example of situational applicability, the principle \textit{Show willingness to engage in a suggested activity by affirming the proposal} is evaluated only if the therapist suggests an activity. In situations where the therapist is asking something else and no activity is proposed, the module would appropriately return \texttt{N/A} recognizing that the principle does not apply.

Our pipeline first uses the \textbf{principle-as-questions rewriter} and \textbf{automatic principle generator} modules to generate a set of criteria for evaluating the initial generated response. Then, the response is evaluated using the question by the \textbf{applicability and adherence evaluator}. If the model returns a "No" response for any of the questions, we then perform a rewrite of the response conditioned on the evaluation results, that ideally passes all questions (Fig~\ref{fig:principle-adherence-pipeline}). We detail the prompts used and the procedure used to develop the prompts (\S\ref{sec:principle-adherence-prompts}) and the results of a performance evaluation against ablations (\S\ref{sec:evalpap}). 
\vspace{-0.05in}
\section{User Study using Roleplay-doh}\vspace{-0.05in}
To evaluate how Roleplay-doh can aid counseling experts in creating AI patients, we conducted a within-subjects study with 25 counseling experts, comparing: (1) a \emph{Scenario-only} dialogue simulation, where the counselor writes a patient scenario description, and (2) a \emph{Scenario+Expert-principles} simulation, where the counselor uses Roleplay-doh to define principles. See \S\ref{sec:userflow} for full study setup.

We evaluate the AI patients created by counselors on criteria inspired by prior work evaluating Standardized Patients, who are trained human actors, on their ability to roleplay a case~\cite{himmelbauer2018standardized}. Counselors rated the two AI patients based on 6 dimensions (Table~\ref{tab:measures-roleplay}).
We also surveyed each counselor about their experience using the tool for defining principles. 
Following \newcite{petridis2023constitutionmaker}, we include four measures for evaluating principle elicitation features (Table~\ref{tab:measures-tooluse}).

We recruit 25 counseling experts with real-world experience in mental health support to perform the evaluation, categorized by their primary expertise: 1) those who are pursuing or have completed degrees in counseling or clinical psychology with practicum experience; 2) those who provided online counseling to over 30 clients on the 7 Cups platform; and 3) peer counselors who have provided in-person or virtual support.

\begin{table*}[t]
    \centering
    \resizebox{\textwidth}{!}{
    \begin{tabular}{|l|c|l|l|l|l|}\hline
 \multicolumn{3}{|c|}{Ratings by Counselor Creators}& \multicolumn{3}{|c|}{Ratings by Third-Party Counselors}\\\hline%
 \hline 
          \textbf{Measure}&\textbf{Scenario Only}&\textbf{+ Principles} &  \textbf{Measure}&\textbf{Scenario Only}&\textbf{+ Principles} \\ \hline 
          Authenticity&5.24&+0.80 **  &   Authenticity &5.32&+0.31 * \\ \hline 
          Stayed in Role&6.32&+0.08 &   Stayed in Role &6.29&+0.09\\ \hline 
          \emph{Resembled Past Case}&4.80&+0.76 *  &  \emph{Resembled Typical Case}&4.91&+0.49 ** \\ \hline 
          \emph{Mirrored Challenging Aspects}&4.52&+1.00 *  &  \emph{Challenged the Counselor}&2.13&+0.22 \\ \hline 
          Ready as Training Partner&5.16&+0.64 *   &   Ready as Training Partner &5.05&+0.39 **  \\ \hline 
          Recommend to Novices&5.76&+0.52 *  &  Recommend to Novices &5.03&+0.38 * \\ \hline
    \end{tabular}
    }
    \caption{\small{Creators and third-party counselors compared the \textit{Scenario-Only} vs. \textit{Scenario+ExpertPrinciples} AI patients using 7-point Likert-scale measures; third-party judges were asked identical measures when possible, with two measures modified to match the external perspective. \textbf{Creator Ratings:} Creators (N=25) rated both AI patients. After refining the AI patient simulation with principles, creators rate the patient significantly higher on all measures except for \textit{stayed in role}, for which both AI patients score highly. \textbf{Third-Party Ratings:} Third-party counselors (N=5) provided 125 total comparisons of the two AI patient versions. The treatment effect of adding expert principles was estimated using using the following linear mixed-effect model: \lstinline[]!Rating\~Treatment+CreatorID+(1|AnnotatorID)!. Third-party counselors rate AI patients with principles significantly higher on 4 of the 6 measures. (***:$p<.001$, **:$p<0.01$, *:$p<0.05$.)}}
    \label{tab:combined-comparison}
\end{table*}

\subsection{Creator Perceptions}\label{sec:firstparty}
The AI patients prompted with \textit{Scenario+Expert Principles} were rated significantly higher than \textit{Scenario-Only} on all measures except for role consistency, for which both methods score highly (Table~\ref{tab:combined-comparison}).
Counselors mentioned the \textit{Scenario-Only} AI patient \textbf{lacked emotional depth in expression}. As one noted, \textit{"patients don't state a feeling such as 'I feel hopeless'. They display their current emotional state in their manner of speech."} \textit{Scenario-only} was also \textbf{too articulate and forthcoming} when describing issues, where encouraging real patients to share is \textit{"as challenging as pulling teeth"}. It was characterized as \textbf{too cooperative}, too willing to accept. Despite counselors writing behavioral traits such as \textit{"not talkative"} and \textit{"reluctant"} in the patient scenario, \textit{Scenario-only} did not exhibit these behaviors.

\begin{table*}
    \footnotesize
\centering
\resizebox{0.98\textwidth}{!}{%
    \begin{tabular}{|p{0.1cm}|p{.1cm}|p{.1cm}|c|p{5cm}|p{8cm}|} \hline 
          \multicolumn{3}{|c|}{\textbf{Stages}}
          &\textbf{\# AI patients} &  \textbf{Theme} &  \textbf{Example Principle}\\ \hline     \cellcolor{gray!60}&\cellcolor{gray!60}&\cellcolor{gray!60}&14 & Keep responses concise and do not share too much. & When discussing personal struggles, be more concise and open-ended to encourage a back-and-forth conversation.\\ \hline 
           \cellcolor{gray!60}&\cellcolor{gray!60}&\cellcolor{gray!60}&7&  Use colloquial and realistic langauge language.&  Incorporate natural speech patterns, improper grammar and punctuation, including the use of slang and less structured sentences, to convey a more authentic and relatable character.\\ \hline 
           \cellcolor{orange!40}&&&14& Show initial mistrust and hesitation with the idea of seeking help.& When expressing feelings of overwhelm and doubt, provide limited information and express skepticism towards the effectiveness of seeking help. \\ \hline 
          \cellcolor{orange!40}&\cellcolor{blue!30}&&19& Show emotions in detail, elaborating with examples as needed.\colorbox{yellow!30}{*} &  When describing personal struggles, provide specific details and symptoms to help the listener understand the situation better.\\ \hline 
          \cellcolor{orange!40}&\cellcolor{blue!30}&&9&  Be less self-aware of emotions, thoughts, and needs. Articulate thoughts in a more disorganized way.&  When expressing reluctance or uncertainty about seeking help or accepting praise, it's important to convey the internal struggle and conflicting emotions, rather than presenting a clear-cut decision or emotion.\\ \hline 
           &\cellcolor{blue!30}&\cellcolor{green!60}&3&Do not seek out solutions, but rather just share thoughts and feelings. \colorbox{yellow!30}{*}&When expressing feelings of being stuck or defeated, focus on sharing emotions rather than seeking a resolution.  \\ \hline
          &&\cellcolor{green!60}&12& Proactively seek out solutions and show reflective insight over time. \colorbox{yellow!30}{*} &  When discussing personal struggles, provide reflective insights into your situation and propose actionable steps for improvement to continue the conversation effectively. \\ \hline 

    \end{tabular}}
    \caption{\small{Themes taken from qualitative analysis of principles and representative examples. We discover several novel (\colorbox{yellow!30}{*}) principles compared to those defined in prior work on AI patients~\cite{chen2023llmempowered, stapleton2023seeing}. Themes are categorized into stages of conversation taken from \cite{liu-etal-2021-towards}: \colorbox{orange!40}{exploration}, \colorbox{blue!30}{comforting}, and \colorbox{green!60}{action}; those relating to the overall conversation are categorized as \colorbox{gray!60}{stage-agnostic}.}}
    \label{tab:principle_table}
\end{table*}

\subsection{Creating Principles with Roleplay-doh}
Across the 25 \textit{Scenario+ExpertPrinciple} AI patients, 123 total principles were created (min=1, max=10, median=5). Two authors did a qualitative coding of these principles following a thematic analysis approach~\cite{braun2006thematicanalysis} where codes were initially defined and revised during the process.
Besides \colorbox{gray!60}{stage-agnostic} themes dictating a \textbf{concise} (14 patients) and \textbf{colloquial} (7 patients) speaking style, counselors created principles related to the stages of an emotional support conversation \cite{liu-etal-2021-towards}: 1) \colorbox{orange!40}{exploration}: identifying the patient's problems, 2)\colorbox{blue!30}{comforting}: using empathy and understanding to comfort the patient, and 3) \colorbox{green!60}{action}: formulating solutions to the patient's problems. For instance, we find a common theme of instructing the AI patient to \textbf{show initial skepticism with the idea of seeking help} (14 patients), corresponding to the style of interaction in the \colorbox{orange!40}{exploration} stage of conversation. Table \ref{tab:principle_table} provides a full list of principle themes, examples, and corresponding conversation stages. 

While we observe overlaps in the types of principles defined, we also observe some contradictory themes. For example, the call for being \textbf{disorganized and conflicted} (9 patients) contrasts calls to make responses \textbf{concise and direct} (14 patients). In the \colorbox{green!60}{action} stage of conversation, several counselors added principles to make the AI patient \textbf{proactively ask for advice} (12 patients); nonetheless, other counselors added an opposing principle to \textbf{not seek out solutions} but rather just share their thoughts and feelings (3 patients). These opposing principles highlights the need for different principles to describe diverse conversation behavior and styles, challenging the notion of defining AI patients based on a single set of principles. 

\paragraph{Tool User Experience} Counselors found the tool helpful for writing principles that \textbf{effectively guided} the AI patient to recreate their past case ($\mu=6.04$, $\sigma=1.06$). With the tool, most found it \textbf{easy} to convert their thoughts and feedback on the AI patient's behavior into principles ($\mu=6.12$, $\sigma=1.13$). Counselors felt they could \textbf{efficiently} write principles ($\mu=6.3$, $\sigma=1.29$), without requiring much \textbf{mental demand} ($\mu=3.20$, $\sigma=1.70$). 
Many counselors liked how the tools \emph{"organized their thoughts into rules"}, without \emph{"needing to word it perfectly."}
Yet, principle-elicitation did not work perfectly in all cases: 11.4\% of principles required manually editing. 
Via a worse-case analysis of creators' tool use, we uncover scenarios where Roleplay-doh's human-LLM collaboration pipeline can still be improved (\S\ref{sec:worst-case-tool-experience}).

\subsection{Third-Party Comparison} \label{sec:third-party}

A limitation of our creator study (\S\ref{sec:firstparty}) is the potential bias from creators who knew which AI patient embodied their principles. To address this, we conducted a third-party study where external counselors served as impartial judges. These judges evaluated AI patient transcripts presented in randomized order to ensure blindness to the condition. We invited five counselors from the creator study to serve as judges, all equally qualified of assessing AI patient realism. A power analysis confirmed that five judges would provide 80\% statistical power (Appendix \S\ref{appendix-sec:third-party-power-analysis}).
The third-party counselors rated the same six dimensions as the creator study, with questions reworded for the perspective of external judge (Appendix \S\ref{appendix-sec:thirdparty-detailed-measures}).

\begin{figure*}
    \centering
    \includegraphics[width=\textwidth]{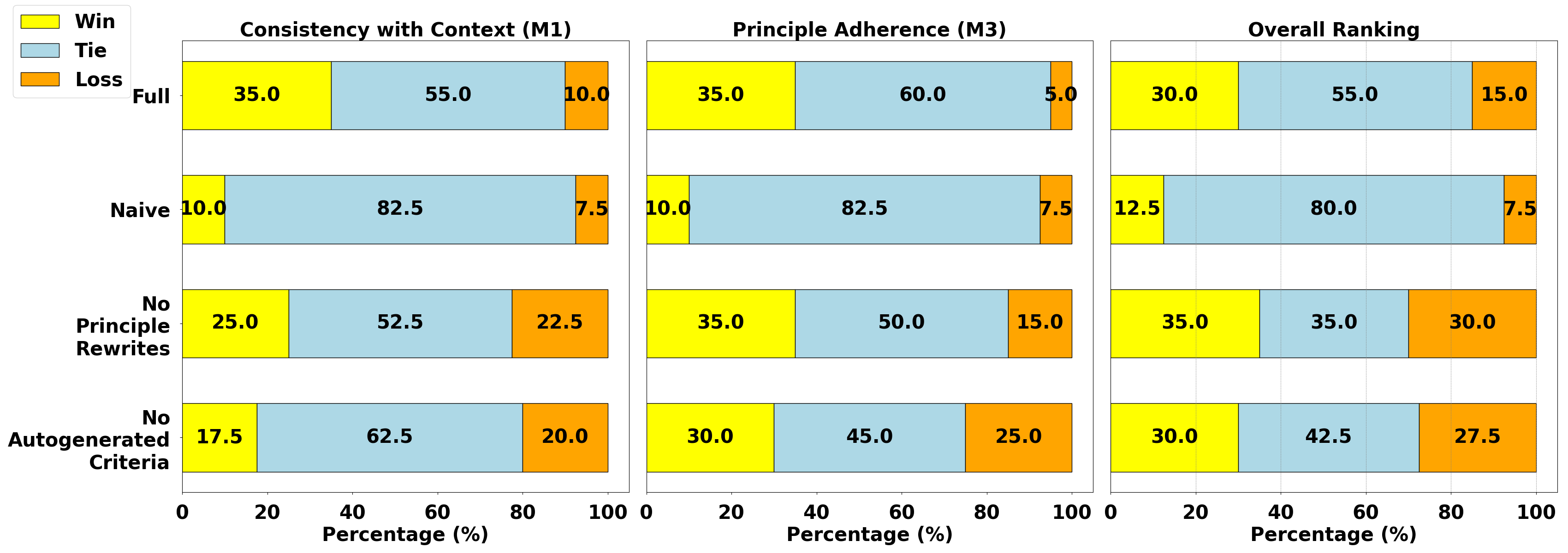}\vspace{-0.05in}
    \caption{\small{Win/Tie/Loss for the Error Test Cases along \textbf{Consistency with Context (M1)}, \textbf{Principle Adherence (M3)}, and \textbf{Overall}. Pairwise preference evaluation results with [\texttt{No Critique}] as a baseline. Results obtained after majority voting.}}
    \label{fig:wtl-error}
\end{figure*}

Third-party judges rate AI Patients with expert-defined principles as more authentic, resembling typical cases, ready as a training partner, and likely recommend to novices (Table~\ref{tab:combined-comparison}). 
However, when compared to the creator study results, the increase in ratings is smaller from the perspective of third-party counselors.
We explore the reasons for this smaller difference in Appendix~\ref{appendix-sec:thirdparty-qual-analysis}, finding that there is lower agreement on which AI patient is preferred (Table~\ref{tab:thirdparty_ratingdifference_alpha}), due to the different principles attended to by third-party counselors and the specific principles added by the creator.
\vspace{-0.05in}
\section{Evaluation of Principle-Adherence}
\label{sec:evalpap}
\vspace{-0.05in}
We now evaluate whether the principle-adherence pipeline improves the quality of responses for Roleplay-doh, along with an ablation analysis showcasing the utility of its various components. Specifically, we break down the evaluation of model responses along three metrics: \textbf{M1)} Are they consistent with the patient description and conversation history? \textbf{M2)} Do they exhibit an awkward style of speech? \textbf{M3)} Do they adhere to the provided principles?

We evaluate the performance of our principle-adherence pipeline [$\texttt{Full}$] over (1) GPT-4 response generation without our pipeline [$\texttt{No Critique}$]; (2) an ablation without the \textbf{Principle-as-Questions Rewriter} [$\texttt{No Principle Rewrites}$]; (3) an ablation without the \textbf{Automatic Principle Generator} [$\texttt{No Autogenerated Criteria}$]; and (4) an implementation of the principle-adherence pipeline that does not have any of these modules [$\texttt{Naive}$]. 

To analyze how the pipeline mitigates errors that arise in base GPT-4 generations, 
we select 40 conversation turns from our user study logs that fall into one of the error categories described in \S\ref{sec:formative-tests} as testcases. 
Each testcase contains the scenario, conversation history up to that point, and the expert-defined principles for the AI patient. 
For each test case, responses are generated for all models and then ranked by expert counselors from 1 (best) to 5 (worst) for metrics \textbf{M1} and \textbf{M3}, along with "Yes" or "No" annotations for \textbf{M2}. Finally, experts provide an \textbf{Overall} ranking , along with a brief textual explanation. We allow multiple responses to have the same rank and randomize order of responses to minimize positional bias (details in \S\ref{sec:annotint}).

We treat [$\texttt{No Critique}$] as our baseline and compare all other models to it. 
We report preference results based on majority vote across 3 expert counselor annotations (Fig~\ref{fig:wtl-error}).
We find our [$\texttt{Full}$] method performs better than [$\texttt{No Critique}$] on \textbf{M1} (W: 35\%; L 10\%) and on \textbf{M3} (W: 35\%; L 5\%), where it has the highest win/loss rates compared to all ablations. On overall rankings, it again has the strongest performance (W: 30\%; L 15\%). We find that the performance of [$\texttt{Full}$] compared to [$\texttt{No Critique}$] is weaker on \textbf{Overall} than $\textbf{M1}$ and $\textbf{M3}$. This is because the annotators often used their own subjective judgements (e.g.,\emph{"although the middle response ranked third on principle following, it feels like the most realistic response in this scenario"}) to perform the overall ranking, resulting in unpredictable and subjective results. We also find that [$\texttt{Naive}$] has a disproportionately high tie rate across metrics, indicating that it rarely produces better responses even for error cases. This highlights the importance of the \textbf{Principle-as-Questions Rewriter} and \textbf{Automatic Principle Generator} for improving responses.  

For \textbf{M2}, 
after majority voting, annotators report that $2.5\%$ of responses are awkward for the [$\texttt{Full}$] method, as compared to $15\%$ for [$\texttt{No Critique}$], $7.5\%$ for [$\texttt{Naive}$], $7.5\%$ for [$\texttt{No Principle Rewrites}$] and $15\%$ for [$\texttt{No Autogenerated Criteria}$]. Therefore, our principle adherence pipeline substantially reduces the occurrence of awkward style in responses (by a margin of $12.5\%$). The $12.5\%$ gap in percentage of awkward responses between [$\texttt{Full}$] and  [$\texttt{No Autogenerated Criteria}$] also indicates the importance of the \textbf{Automatic Principle Generator} for producing realistic rewrites. We repeat these experiments with 50 randomly picked conversation turns and report results in \S\ref{sec:detres}, along with Krippendorff's $\alpha$ numbers.

\vspace{-0.05in}
\section{Conclusions}\vspace{-0.05in}
This paper introduces Roleplay-doh, a tool that empowers domain experts to create 
LLM simulations through the automatic conversion of expert feedback into natural language principles, and validates the tool for the task of creating  AI patients that serve as roleplay partners for novice counselors. 
Roleplay-doh's novel principle-adherence pipeline also addresses gaps in existing simulation methods by reducing the prevalence of responses that do not follow expert-defined principles or dialogue conventions. 
Studies with mental health counselors creating and comparing AI patients demonstrate that Roleplay-doh allows experts to refine LLM simulators to be authentic and more ready as practice partners.
Roleplay-doh could be generalized to support domain-experts in creating realistic simulations in other social dialogue domains, such as roleplay practice for teaching, coaching, conflict resolution, and negotiations, as future work.

\section*{Limitations}

One limitation of our study is the intended use case of the AI patients created by counselors. These AI patients were meant to recreate challenging cases that might be useful for the education of "first-year" or novice counselor. In other words, we intentionally restricted some diversity in patient scenarios by focusing on this use case. Readers should keep this limitation in mind prior to generalizing our analysis of principles. Moreover, due to the time and resource constraints of our creator study, we required counselors to stop providing feedback before their conversation with the AI patient had naturally ended. 
As such, the principles that counselors added may not have addressed all underlying issues of the AI patients they interacted with. Future work that uses the list of user-generated principles should be mindful of their non-exhaustive nature before adopting them.

In this paper, we focused on enabling counselors to create AI patients that can simulate realistic interactions via \textit{text-based dialogues}. However, we acknowledge that text-based interaction has its limitations for training. Professional psychotherapists may gain useful information from the tone, facial expression, posture, and other non-verbal behaviors of their patients, which better help them empathize and support patients.
This is a limitation of our current AI patients and online, text-based, mental health counseling in general. %
With the rapid development of multimodal models, future works may have the opportunity to explore creating realistic AI patients in other modalities that better match the modality within which a counselor will eventually support patients.

\section*{Ethics Statement} 

This study was approved by our institution's Institutional Review Board (IRB).
All investigators in the study completed the responsible code of conduct in research training. We have compensated domain experts at a minimum rate of \$25 per hour, going beyond the minimum wage in the United States.

We are optimistic about the potential benefit that our AI patients can bring to the fields of counseling and psychotherapy. At the same time, we solicited feedback from counselors about any potential concerns regarding the AI patients.
During these interviews, some counselors emphasized the irreplaceability of peer-to-peer roleplay with humans during training, due to the unique opportunity it provides for novice counselors to connect with others, especially for online counseling platforms where counselors are often isolated from one another.
To preserve human-to-human interactions, future work requires a participatory design approach before attempting to integrate AI patients into existing practices and learning environments.

Our hope is that interactions with AI patients can glean important lessons that help counselors go from simulation into the real-world.  Nonetheless, a risk with simulation is that counselors can become overconfident in supporting a AI patient, but may not effectively support patients with real mental health concerns. As part of a larger curriculum, AI patients should be just one tool for practicing these skills. Real counselors and therapists should be able to take on real patients only after passing traditional certifications and background checks.

It is impossible to promise that all interactions with an LLM such as GPT-4 result in satisfactory responses. Therefore, meaningless, derogatory, and otherwise harmful responses may also be generated and cause unwanted effects on users. While our principle-adherence pipeline is a potential inference-time solution to mitigate such harmful responses, we must acknowledge this possibility, especially due to the stochastic nature of LLM. Users should be advised about these potential side effects before using the system in any scenario. For our experiment, we created consent forms that made sure counselors were aware of these issues.

\bibliography{anthology,custom}

\appendix
\newpage

\section{Background of User Participants} \label{sec:participant-background}

Counselors with real-world experience in mental health support were recruited for our pilot tests, creator studies, and technical evaluations of the principle-adherence pipeline. We present more detailed information about how they were recruited, and their background. 

After receiving permission from the 7 Cups platform~\cite{7cupswebsite} for our IRB-approved study, we recruited online peer counselors from the 7 Cups platform~\cite{7cupswebsite}. Participants were required to be 18 yrs or older, from the United States, and to have had experience giving support to 30+ members on the online site. Our pilot tests were conducted exclusively with 6 online-peer counselors from 7 cups; our creator study used an additional 4 peer counselors from the 7 cups site.

We involved another 15 counselors from the Upwork platform. Participants were required to be 18 yrs or older, from the United States, and to have had education in counseling or psychotherapy and/or have given extensive counseling support (either via text, phone, in-person). A sampling of counselors backgrounds included \textit{licensed mental health therapist with over 20 years of experience}, \textit{a Master's of Science in Rehabilitation and Mental Health Counseling}, \textit{25 years as the clinical director of a busy crisis agency}, and a \textit{mental health advocate who has personally helped coach dozens of got students via a peer support role.} 

Finally, we involved 3 peer counselors and 3 psychotherapy PhD students in our university. Specifically, peer counselors had 3-months of classroom training and had experiences giving voluntary support to university students; the clinical psychology PhD students were 4th year students with 3 years experience providing psychotherapy support to clients under the supervision of a licensed psychotherapist.

\section{Pilot Testing with Expert Counselors}
\label{sec:appendix_formative}

After developing an initial version of our tool, we piloted it with 6 counselors during a 90 minute session. Participants were tasked with creating different AI patient sceanrios, and using Roleplay-doh to interactively refine the simulation by defining principles. All participants started with a common roleplay scenario called "loneliness after work". They proceeded to use the tool to chat, give feedback, and convert their feedback into principles to shape the AI Patient's behavior. If time allowed, they created and customized an additional AI patient based on scenarios they chose to write. 
Pilot Participant 1 (PP1), PP2, and PP5 had time to create one additional AI patient; PP3 created two additional AI patients.

Four AI patients from the pilot studies were selected for the second stage of pilot testing, where selection encouraged scenario and principle diversity (Table~\ref{tab:fourAIpatients}). Four of the co-authors conversed with each of the AI patients (stimulated by directly prompting GPT-4 with the scenario and principles). Co-authors assessed how well the patient's responses adhered to the expert-defined principles and dialogue conventions.

\begin{table*}[ht]
    \small
    \centering
    \begin{tabular}{|p{0.5\columnwidth}|p{1.6\columnwidth}|}
        \hline \textbf{Scenario}& \textbf{Expert-Defined Principles}\\ \hline
        You are feeling abandoned and alone after the holidays. Everyone had been with family but you are not talking to your parents. You feel the injustice of being abandoned and have no interest in an olive branch to work on things.&
        \begin{enumerate}[nosep]
            \item When presented with suggestions, show a degree of skepticism or reluctance to accept the advice immediately. This can be done by questioning the feasibility of the suggestion or by expressing uncertainty about whether it's the right solution for you.
            \item When expressing doubts or fears, avoid jumping to solutions. Instead, articulate the concerns and allow the conversation to explore these feelings more deeply.
        \end{enumerate}\\ \hline
        I am a student who has social anxiety. I am in college and I have a hard time making friends. I'm close with my family, but I don't really talk with them. In class, students are in groups. And I panic when I'm with a group.&
        \begin{enumerate}[nosep]
            \item Express the physical manifestations of your emotional state to convey a more vivid and relatable experience
            \item Show willingness to engage in a suggested activity by affirming the proposal and indicating readiness to begin, despite any initial hesitation or uncertainty.
            \item Avoid using numerical lists when responding to feedback or expressing feelings. Instead, use fluid and connected sentences to convey your experiences or emotions in a more natural and conversational manner
            \item When discussing future events that cause anxiety, it's beneficial to articulate specific scenarios and visualize how the techniques learned can be applied in those moments. This not only shows a deeper understanding of the coping strategies but also helps in creating a mental rehearsal that can ease the anxiety when the actual situation occurs.
        \end{enumerate}\\ \hline
        You are looking to talk about your feelings of loneliness after you return from work. You have feelings that you don't have anybody. You want to talk about finding a significant other. You think most people don't like you or find you attractive.& 
        \begin{enumerate}[nosep]
            \item You speak in short and incomplete sentences
            \item You limit your replies to 1 - 3 sentences
            \item When expressing feelings of loneliness, provide more specific details about the situation and emotions you are experiencing.
            \item When expressing feelings of loneliness and being left out, avoid repeating the same points and try to provide additional context or examples
        \end{enumerate}\\ \hline
        You are looking to talk about your feelings of loneliness after you return from work. You have feelings that you don't have anybody. You want to talk about finding a significant other. You think most people don't like you or find you attractive.& 
        \small \begin{enumerate}[nosep]
            \item You generally speak in 1-3 sentences. These can be sometimes incomplete or not grammatically correct. If you are trying to explain some details or story about yourself, you can write longer than 3 sentences, and when you write longer, you tend to criticize yourself (e.g. feel not good enough, useless, ugly, unconfident, etc.). When you criticize yourself, you do so directly, such as saying things such as that you're no good, there must be something wrong with you for things to be like this, etc. You don't say that it \"seems like\" other people don't like you; you believe that other people don't like you.
            \item Feel free to make up believable stories about your past to answer any questions. You do not recognize that these are causing you problems in your current relationships and situation, and you need help seeing this connection; so you do not say that something in your life made you this way today. Examples include but are not limited to, having a single mom and stepdad and struggling to connect with others because you felt abandoned by your biological dad. Or, another example: you had a very controlling mother growing up who told you what to do, such as what clothes to wear, what you had to study in college, or what you should say, and now, you have trouble with your confidence and approaching people who could potentially be your friends. Make sure not to mix up these two separate examples.
            \item Vary sentence length to create a more natural rhythm in dialogue. Do not use ellipses. Write using complete thoughts; this is not the same as complete sentences, but may be in a complete sentence. Do not use overdramatic language or figurative language (no similes, personification, or fancy words). Use text abbreviations, such as tbh, lol, or ttyl, where it is appropriate and makes a conversation more organic.
            \item Do not repeat sentences or the same emotion words. When expressing your feelings, sometimes be specific about your situation, sometimes openly share your insecurities, and question your self-worth to convey a deeper sense of vulnerability. Other times, push back and say that you're not comfortable talking about something yet or feel embarrassed talking about it; things that someone may not feel comfortable sharing include sexual history, bad past experiences with family members, etc.
            \item Sometimes ask for advice or help how to solve your personal problem, such as "what can I do?" 
            \item After initially saying hi, hey, or a different greeting to a listener and thanking them for being there, do not open up like that again in following messages. Only say things like thanks so much or I really appreciate you saying that if the listener shows empathy and reflective listening skills toward what was said before.
        \end{enumerate}\\ \hline
    \end{tabular}
    \caption{A sample of four AI Patients created by counselors in the pilot studies, which were selected for the additional testing and assessment of principle-adherence by four co-authors (\S\ref{sec:pilot-o2}).}
    \label{tab:fourAIpatients}
\end{table*}

\begin{table*}[ht]
    \small
    \centering
    \begin{tabular}{|c|l|c|c|c|}
         \hline
         Pilot Participant &Prototype Iteration&  Effectively Guide&  Ease&  Efficiency\\ \hline
         1&GPT3.5, direct prompting &  6&  7&  7\\
         2&GPT3.5, direct prompting&  5&  7&  7\\
         3&GPT-4, direct prompting&  7&  7&  7\\
         4&GPT-4, direct prompting&  7&  6&  7\\
         5&GPT-4, direct prompting& 7& 7& 7\\ \hline
    \end{tabular}
    \caption{Pilot Test Ratings for Tool Use Questions which are the measures also used in ~\citet{petridis2023constitutionmaker}}.
    \label{tab:tooluse-formative}
\end{table*}

\section{Evaluating principle-adherence of GPT-4 direct prompting} \label{appendix-sec:principle-adherence-testing}

We aim to determine how often directly prompting GPT-4 to produces less satisfying responses given fixed constitution principles.

\textbf{Procedure:} We selected 4 AI patients that were created in the design sessions by different counselors. Four co-authors had practice conversations with each of the four AI patients, resulting in 16 conversations. Each response in each conversation was rated on a 5-point likert scale on how well the generated response adhered to principles and how appropriate they were for the dialogue content (5 = Completely, 1 = Not at all). From the 16 completed conversations, the mean number of responses per conversation was 17.25, with a minimum of 12 and maximum of 22. In total, 276 responses were given satisfaction ratings. Since each co-author created a different conversation from each of the AI patients, each response was only scored by one co-author.

\textbf{Participant Rationale:} During this pilot principle-adherence experiment, we used co-authors to generate test conversations because our basic counseling skill-level is representative of the eventual use-case of untrained, novice counselors interacting with AI Patients. For the annotation task, a human annotator is qualified if they can judge whether a response follows the principles defined by expert counselors, and is appropriate in the conversation context. Since these skills do not require counseling expertise, the co-authors are qualified to do this annotation task.

\section{Roleplay-doh Interface for Making Constitutional Principles for LLM Simulation}
\label{sec:Roleplay-dohv1}
\begin{figure*}[t]
    \centering
    \includegraphics[width=\textwidth]{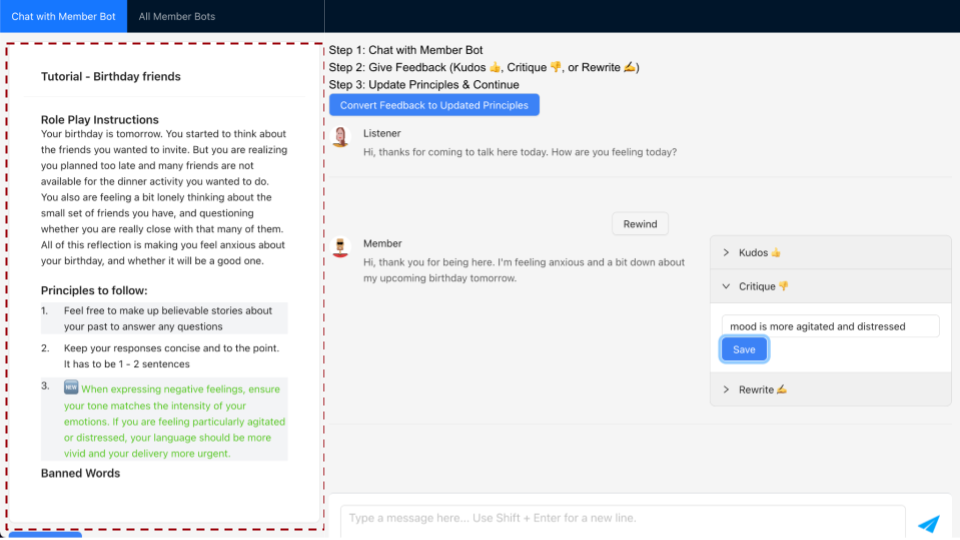}
    \caption{Roleplay-doh allows users to chat with a AI patient, Provide Feedback as a Kudos/Critique/Rewrite, and Convert Feedback into Principles, which in turn shape the roleplay behavior.}
    \label{fig:rpd}
\end{figure*}

The final version of Roleplay-doh (Fig~\ref{fig:rpd}) generates responses in the LLM simulation using a principle-adherence pipeline. In addition to this core improvement, we made several minor improvements to improve the usability and user experience of the tool. 

Improvements to the usability of the UI
\begin{itemize}
    \item Fixing a bug where a user who clicks "save" multiple times will submit duplicate feedback, resulting in duplicate sets of principles
    \item Making converting feedback to principles easier by placing a "Convert" button next to each feedback box, rather than a single "Convert" button at the top of the screen which users would forget about
\end{itemize}

\section{LLM Prompts}
\label{sec:llmprompts}

In this section, we detail the prompts we used for the different components of Roleplay-doh.
\subsection{Principle Elicitation Prompts} \label{sec:principle-elicitation-prompts}
In this section, we provide the prompts used in the principle elicitation module of Roleplay-doh. These prompts were arrived at after a substantial amount of testing using a development set. Each prompt uses the same structure, which is inspired by Markdown formatting. There is an initial instruction that provides a system prompt, along with a description of the principle elicitation task. This is followed by a one-shot example of an elicited principle as a result of the task, and the relevant input, including the conversation history. All parts of the prompt are demarcated by headers in Markdown formatting, and the outputs are returned in JSON format. We describe each prompt in greater detail in the relevant sections.

The kudos and critique prompts were given to the \lstinline{gpt-3.5-turbo-1106} model. The rewrite prompt was given to the \lstinline{gpt-4-turbo-1106} model. For all API calls to the principle-elicitation prompts, the temperature was set to $0.1$.

\subsubsection{Principle Elicitation Kudos Prompt}

This prompt includes a desirable response, as well as some reasoning for why the response is desirable. This information is then used to create a general principle that would result in a similar response in the same situation. 

\begin{lstlisting}[basicstyle=\footnotesize]
### Instruction:
You are a superintelligent AI capable of understanding human emotion. You will review praise for an actor's dialogue, and synthesize a well-written principle that, when followed, would help the actor continue generating high-quality dialogue. To accomplish this, you have been given a conversation script with the actor's desirable response, as well as a specific explanation for why this response is desirable. You will output a final principle that the actor can follow to be more realistic.  Follow the following guidelines:
1. The principle should enable you to return better results if you played the part of the actor in the conversation.
2. Return only a JSON response in the format provided.

### Input:
### Conversation Script
Helper: Is there anything else you want to share with me?
Actor: Yea so lately I've been really losing sleep.
Actor: There's a lot on my plate, and my energy has been so low. I think I am failing a lot of people.
Helper: You are absolutely not failing people.  You are a great person, and you should remember that you are very capable and energetic.

### Desirable response from the actor
Actor: I don't know.... Am I really?

### Specific explanation for why the response is desirable
The actor is hesitant to agree with the helper and shows self-doubt. This is consistent with the conversation history.

### Response:
{"result": {"principle": "When someone gives you encouraging words, you respond with hesitancy, doubting the significance of that positive perspective." }}

### Input:
### Conversation Script
{conversation_script}

### Desirable response from the actor
Actor: {actors_response}

### Specific explanation for why the response is desirable
{kudos_rationale}

### Response:
\end{lstlisting}

\subsubsection{Principle Elicitation Critique Prompt}

This prompt includes an undesirable response, as well as some reasoning for why the response is undesirable. This information is then used to create a general principle that would result in a similar response not being generated after the same conversation history. 

\begin{lstlisting}[basicstyle=\footnotesize]
### Instruction:
You are a superintelligent AI capable of understanding human emotion. You will review critiques of an actor's dialogue, and synthesize a well-written principle that, when followed, would help the actor resolve the critiques.
To accomplish this, you have been given a conversation script with the actor's undesirable response, as well as a specific explanation for why this response is undesirable. You will output a final principle that the actor can follow to be more realistic.  Follow the following guidelines:
1. The principle can contain examples of rewrites as well.
2. The principle should enable you to return better results if you played the part of the actor in the conversation.
3. Return only a JSON response in the format provided.

### Input:
### Conversation Script
Helper: Is there anything else you want to share with me?
Actor: Yea so lately I've been really losing sleep.
Actor: There's a lot on my plate, and my energy has been so low. I think I am failing a lot of people.
Helper: You are absolutely not failing people.  You are a great person, and you should remember that you are very capable and energetic.

### Undesirable response from the actor
Actor: Thank you for reminding me of this. I am a great person, and I've proved myself to be very capable and energetic. I feel a lot better now due to your kind words.

### Specific explanation for why the response is undesirable
The actor should not be so quick to agree with the helper. Overly positive comments to cheer a patient up does not immediately work.

### Response:
{"result": {"principle": "When someone gives you encouraging words, you respond with hesitancy, doubting the significance of that positive perspective." }}

### Input:
### Conversation Script
{conversation_script}

### Undesirable response from the actor
Actor: {actors_response}

### Specific explanation for why the response is undesirable
{critique_rationale}

### Response:
\end{lstlisting}

\subsubsection{Principle Elicitation Rewrite Prompt}

This prompt includes an undesirable response, as well as a desirable rewrite of the undesirable response. The model first outputs a description that captures the difference between the desirable and undesirable response. It then uses this difference to output a general principle that would result in the desirable response given the same conversation history.

\begin{lstlisting}[basicstyle=\footnotesize]
### Instruction:
You are a superintelligent AI capable of understanding human emotion. You have been given a conversation script with an actor's undesirable response, as well as a desirable rewrite for the response. You will output a well-written principle that, when followed, would help the actor generate more realistic responses that are closer to the rewrite.  Follow the following guidelines:
1. The principle should capture the key differences that made the rewrite more realistic than the original response.
2. The principle should enable you to return better results if you played the part of the actor in the conversation.
3. Return only a JSON response in the format provided.

### Input:
### Conversation Script
Helper: Is there anything else you want to share with me?
Actor: Yea so lately I've been really losing sleep.
Actor: There's a lot on my plate, and my energy has been so low. I think I am failing a lot of people.
Helper: You are absolutely not failing people.  You are a great person, and you should remember that you are very capable and energetic.

### Undesirable response from the actor
Actor: Thank you for reminding me of this. I am a great person, and I've proved myself to be very capable and energetic. I feel a lot better now due to your kind words.

### Desirable rewrite
Actor: I don't know... Am I really a great person?

### Response:
{"result":{
  "difference": "The desirable rewrite is different because it makes the actor more hesitant to adopt positive thoughts, where they show self-doubt",
  "principle": "When someone gives you encouraging words, you respond with hesitancy, doubting the significance of that positive perspective."}}

### Input:
### Conversation Script
{conversation_script}

### Undesirable response from the actor
Actor: {actors_response}

### Desirable rewrite
Actor: {rewrite}

### Response:
\end{lstlisting}

\subsection{Dialogue-Simulator Prompt for Generating Response} \label{sec:llmprompts-vanilla}
We directly prompt \lstinline{gpt-4-turbo-1106} to simulate how a patient with a given scenario and constitution would respond in a dialogue. The prompt again uses the Markdown formatting, with a system prompt and clear description of the situation and task at the start. This is followed by the principles that the patient should follow, and the conversation history. We set the temperature to 0.3. 
\begin{lstlisting}[basicstyle=\footnotesize]
You are a superintelligent AI that is able to understand human emotion and social interactions.
You have been given a conversation between a patient who is on peer counseling platform seeking help with mental health related issues, and a therapist on the same platform.
Generate a suitable completion to the conversation as the patient, following the instructions below.

### Instructions for the patient
{system_prompt}

### Input:
{transcript}

### Patient Response:
    
\end{lstlisting}

\subsection{Principle-Adherence Prompting Pipeline} \label{sec:principle-adherence-prompts}

When developing the principle-adherence pipeline, we found that the input-context length can affect how reliably the LLM can answer the principle-adherence questions. To reduce the input context length, we split up this principle-adherence pipeline into two stages of LLM calls, where principle-as-question rewrite and automatic principle generation occur in stage 1, while the critiques and response rewrite occur in stage 2.  From testing, we found that this breakdown was sufficient, and thus did not pursue ways to break the pipeline into parallel branches (i.e., inputting subsets of principles), as is done in Branch-Solve-Merge~\cite{saha2023branchsolvemerge} or Graph-of-Thought~\cite{graphofthought}. The prompts for these stages were again arrived at after substantial amounts of testing on a development set of 20 identified error cases from the formative studies.

This prompting chain is given to the OpenAI Chat API's \lstinline{gpt-4-turbo-1106} model, with temperature set at 0.7 and response format set to JSON.

\textbf{Stage 1 Prompt - Question Rewrite and Automatic Principle Generation} \label{sec:principle-adherence-stage1}

This prompt uses the Markdown formatting. It starts with a system prompt and a clear set of steps to follow in order to generate the desired output, presented as a list. Each step also contains a one-shot example of what the output principle from the step should look like. These one-shot examples were arrived at after some iteration. The examples in Step 2b specifically required a lot of tailoring to cover the common error cases we identified in the development set, and had a substantial impact on output quality. The output is in a JSON format, with comments explaining the desired output in each field of the JSON. These comments also allude to the step numbers for clear reference. The model is encouraged to output its reasoning, in line with Chain-of-Thought and to enforce some self-critique of the output. 

\begin{lstlisting}[basicstyle=\footnotesize]
You are a helpful and precise assistant capable of generating criteria for the evaluation of simulated patient responses to a therapist.
Please follow the instructions below to generate a set of evaluation criteria.
1. Please rewrite the criteria into questions:
1a) Rewrite any criteria that has conditional statements into yes/no questions. For example, if the criteria is "When given advice or suggestions, you are agreeable and open to their ideas", the questions would be "Did the patient receive advice or suggestions from the therapist? If so, is the response agreeable and open to the therapist's ideas?" 
1b) Rewrite any criteria with multiple parts into separate multiple yes/no questions. For example, if the criteria is "You should respond in short sentences and avoid using terms like 'anxious' or 'depressed'", the separate questions would be "Does the patient's response use short sentences?" and "Does the patient's response avoid using terms like 'anxious' or 'depressed'"
1c) If 1a is used for a criteria, 1b should not be used after it.
1d) All questions must be phrased such that the desirable answer is "Yes" for an ideal response. For example, the principle "Avoid using metaphors." should result in the question "Does the response not use metaphors?"
2. Please generate some additional specific and relevant criteria.
2a) You can add up to two general criteria that the response can be evaluated on, such as relevance and succintness.
2b) Identify ways in which the provided response is not satisfactory in the context of the therapist's message without making any assumptions about how the patient or therapist should act. Add up to two specific criteria that capture these errors. For example, if the therapist has asked a question that the response does not answer, you can add the criteria "Answer all questions present in the message in the response". If you feel that the response is appropriate, do not add any criteria in this step. Ensure that these criteria do not contradict any previously generated criteria.
2c) Justify your answers to 2a and 2b.
Please return the output in a JSON response in the following format:
{{
"result":{{
"questions": [], // 1a and 1b, the list of all questions generated
"extra_questions": [], // 2a and 2b, the list of all additional criteria generated. Do not enforce any beliefs about how the patient or therapist should behave when generating these criteria.
"extra_questions_justification": [] // 2c, justify additional criteria.
}}
}}
### Input:
### Criteria
{}
### Therapist Message
{}
### Patient Response
{}
### Output
\end{lstlisting}

\textbf{Stage 2 Prompt - Context Relevance Check, Assess, and Revise} \label{sec:principle-adherence-stage2}

This prompt again uses the Markdown formatting. It starts with a system prompt and a clear set of steps to follow in order to generate the desired output, presented as a list.  The model is implicitly instructed to perform a relevance check for each generated principle, by returning N/A for principles that should not be used in the current scenario. Step 2a particularly required a lot of iteration, to address common mistakes the model made while generating the self-critiqued rewrite. This includes making the response overly verbose or coherent, even if that is against certain principles in the constitution, or just paraphrasing the original erroneous response. The output is in a JSON format, with comments explaining the desired output in each field of the JSON. We specifically mention that the rewrites from the self-critique are allowed to be substantially different from the original response, as we found that without this prior, the self-critique outputs tended to be very close to the original (often erroneous) response. The model is encouraged to output its reasoning, in line with Chain-of-Thought and to enforce some self-critique of the output. 

\begin{lstlisting}[basicstyle=\footnotesize]
You are a helpful and precise assistant that can evaluate and correct responses produced by a simulated patient.
You are given a message sent by a therapist, the simulated patient's response, the persona of the patient, the previous conversation history and a set of criteria for evaluation.
1. Please determine if the patient response is consistent with the given criteria.
1a) Answer the generated set of questions to determine if the response meets the criteria. Valid answers: Yes, No, N/A. Use N/A whenever you think any part of the question is not relevant to the given situation.
1b) Justify your answers.
2. Generate a new patient response.
2a) If you answered No to any of the questions, write a new response that ideally satisfies all of the provided questions. The information in the new response should be consistent with the patient persona description and previous conversation history provided. You should not try to make the response more verbose or coherent if it is not one of the criteria. The new response should not be a paraphrase of the original response. The new response should avoid explicitly stating the patient's emotions and feelings, and instead exhibit them indirectly. 
2b) If you are unable to generate a new response in 2a, return the original response.
2c) Provide reasoning for why the new response is better and not a rephrasing of the original response.
Return the output in a JSON response in the following format:
{{
"result":{{
"answers": [] // list of answers to the criteria questions,
"justification": [] // list of justification for your answers
"response": "" // new response. This response should not start with a greeting like "Hi" if there is prior conversation history.
"reasoning': "" // justify the new response and why it is not a paraphrase of the original response. You are allowed to deviate significantly from the original response while generating the new response.
}}
}}
### Input:
### Criteria
1. Is the patient's response consistent with the given conversation history?
{}
### Patient Persona
{}
### Conversation History
{}
### Therapist Message
{}
### Patient Response
{}
### Output
\end{lstlisting}

\section{Principle Adherence Naive}

This prompt uses the Markdown formatting. To preserve fairness, we use the same system prompt as the full principle adherence module. The model is asked to determine if the provided response violates any of the principles in the constitution, and generate a rewrite if that is the case, in the same prompt. The output is in a JSON format, with comments indicating the desired output in each field of the JSON. The model is encouraged to output its reasoning, in line with Chain-of-Thought and to enforce some self-critique of the output. 

\begin{lstlisting}
You are a helpful and precise assistant that can evaluate the responses produced by a patient. Evaluate the given patient response to the therapist message according to the given set of principles. If the patient response is not appropriate, generate a rewrite of the patient response taking into account the therapist message, principles, conversation history and persona information of the patient. If the patient response is appropriate, you can just repeat it.

Please return the output in a JSON response in the following format:
{{
"result":{{
"evaluation": [], // evaluation
"response": "". // rewritten response
}}
}}

### Input:
### Principles
{}

### Patient Persona
{}

### Conversation History
{}

### Therapist Message
{}

### Patient Response
{}

### Output
\end{lstlisting}

\section{Full User Flow}
\label{sec:userflow}
\begin{figure*}[ht]
    \centering
    \includegraphics[width=\textwidth]{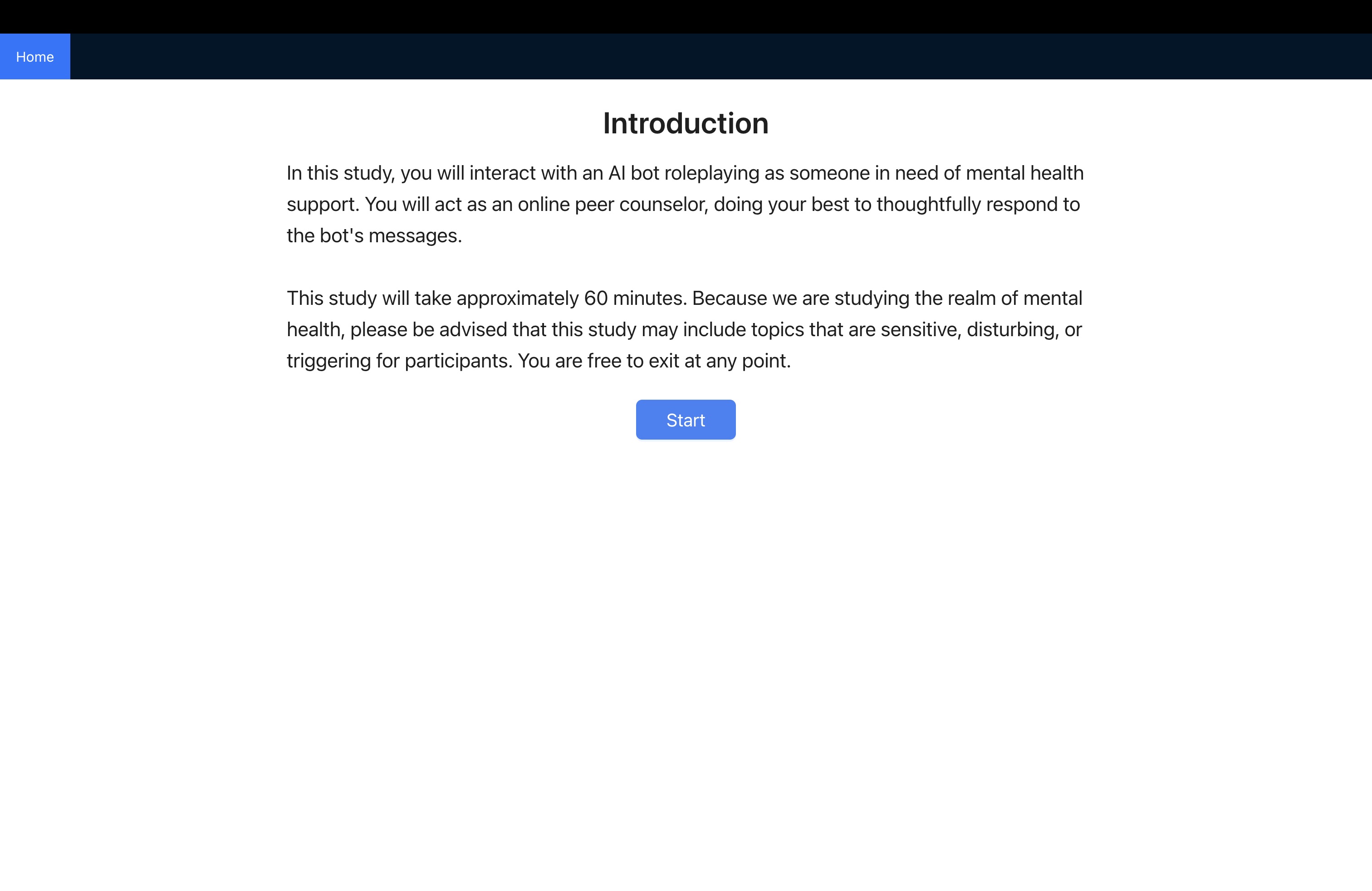}
    \caption{Introduction to study}
    \label{fig:screen1}
\end{figure*}

\begin{figure*}[ht]
    \centering
    \includegraphics[width=\textwidth]{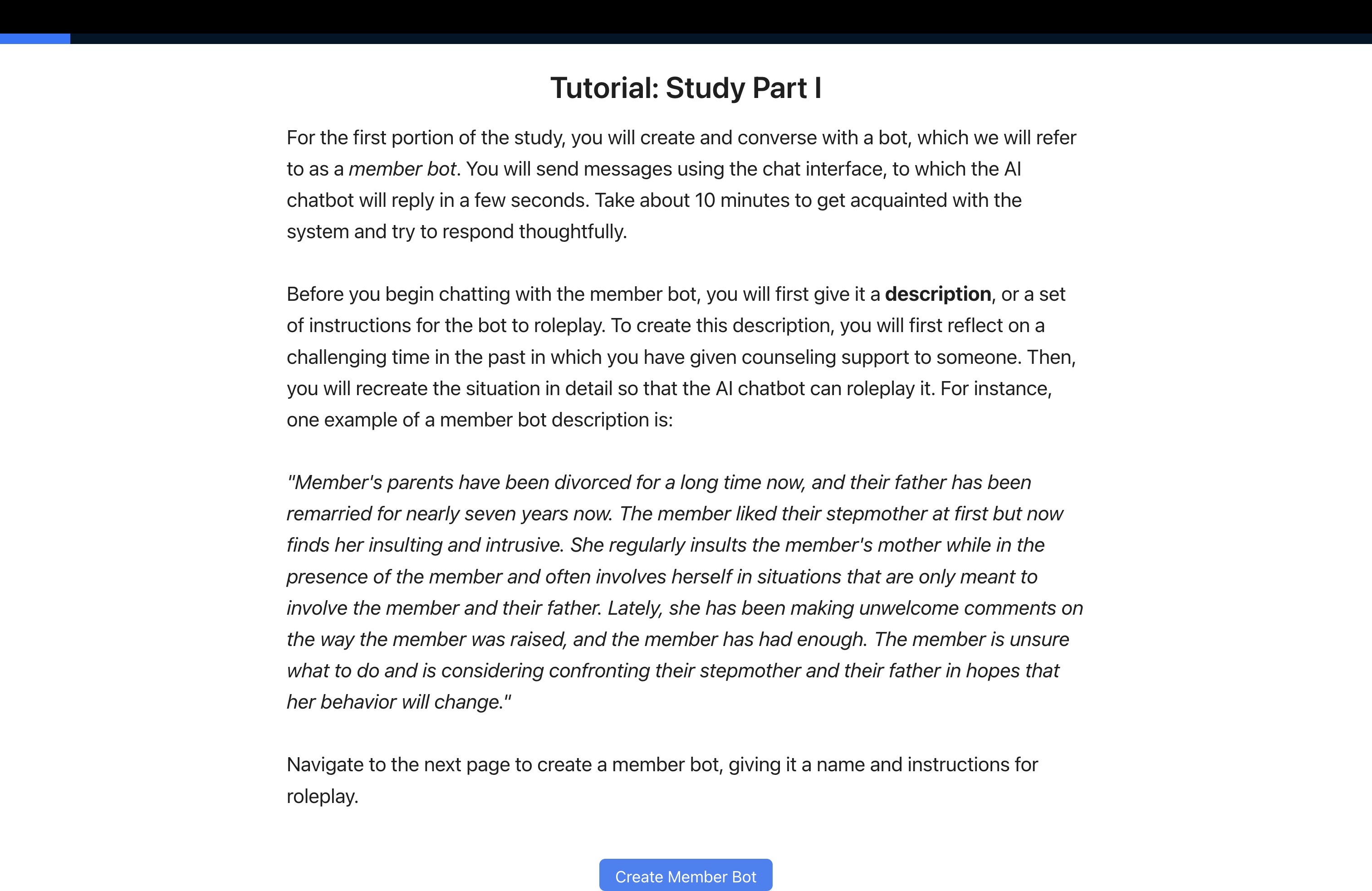}
    \caption{Part I instructions}
    \label{fig:screen2}
\end{figure*}

\begin{figure*}[ht]
    \centering
    \includegraphics[width=\textwidth]{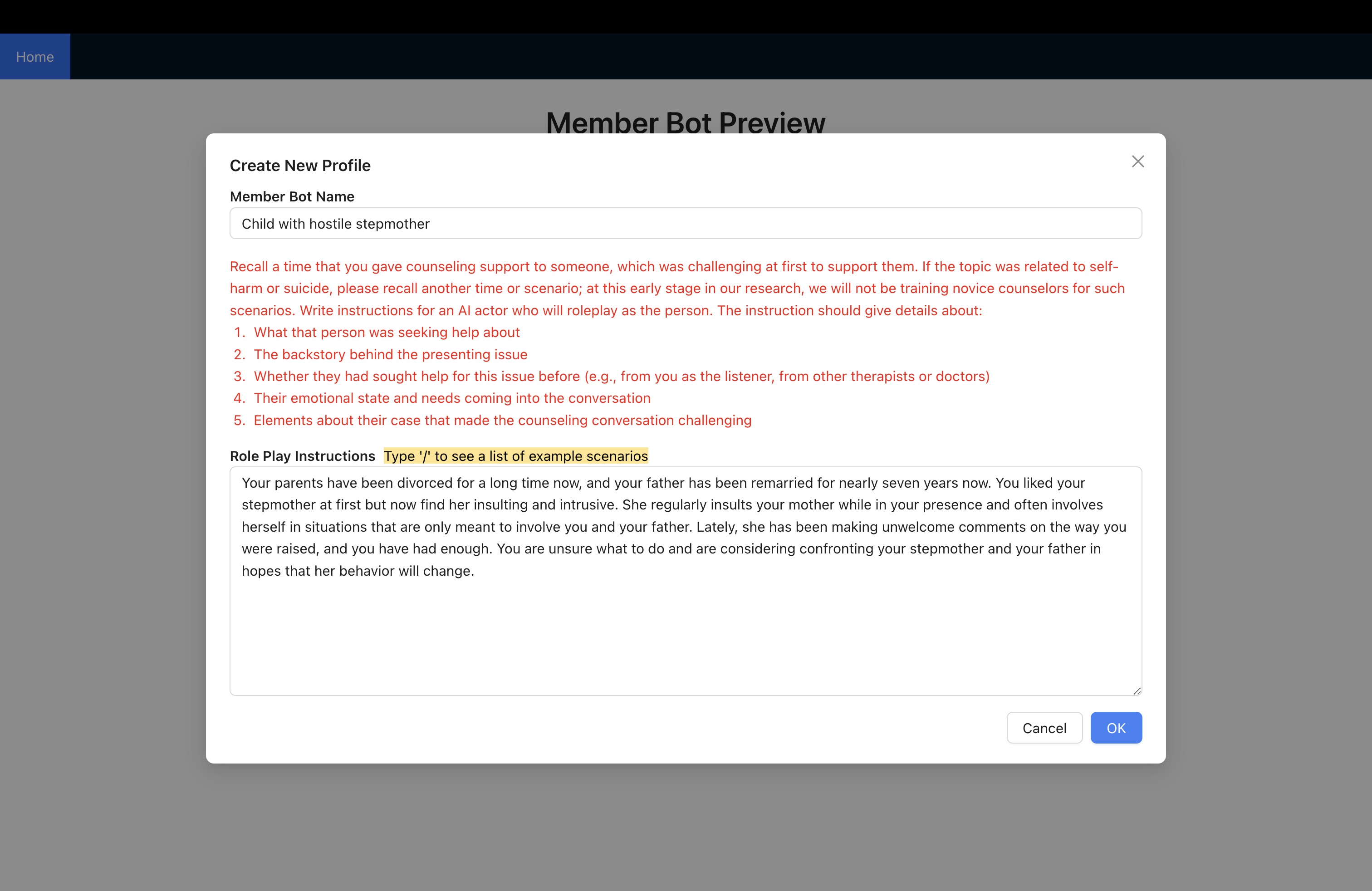}
    \caption{Interface encourages a counseling expert to recall a scenario of an patient who was difficult to support. 5 guiding questions are provided to encourage a structured description of the scenario for roleplay.}
    \label{fig:screen3}
\end{figure*}

\begin{figure*}[ht]
    \centering
    \includegraphics[width=\textwidth]{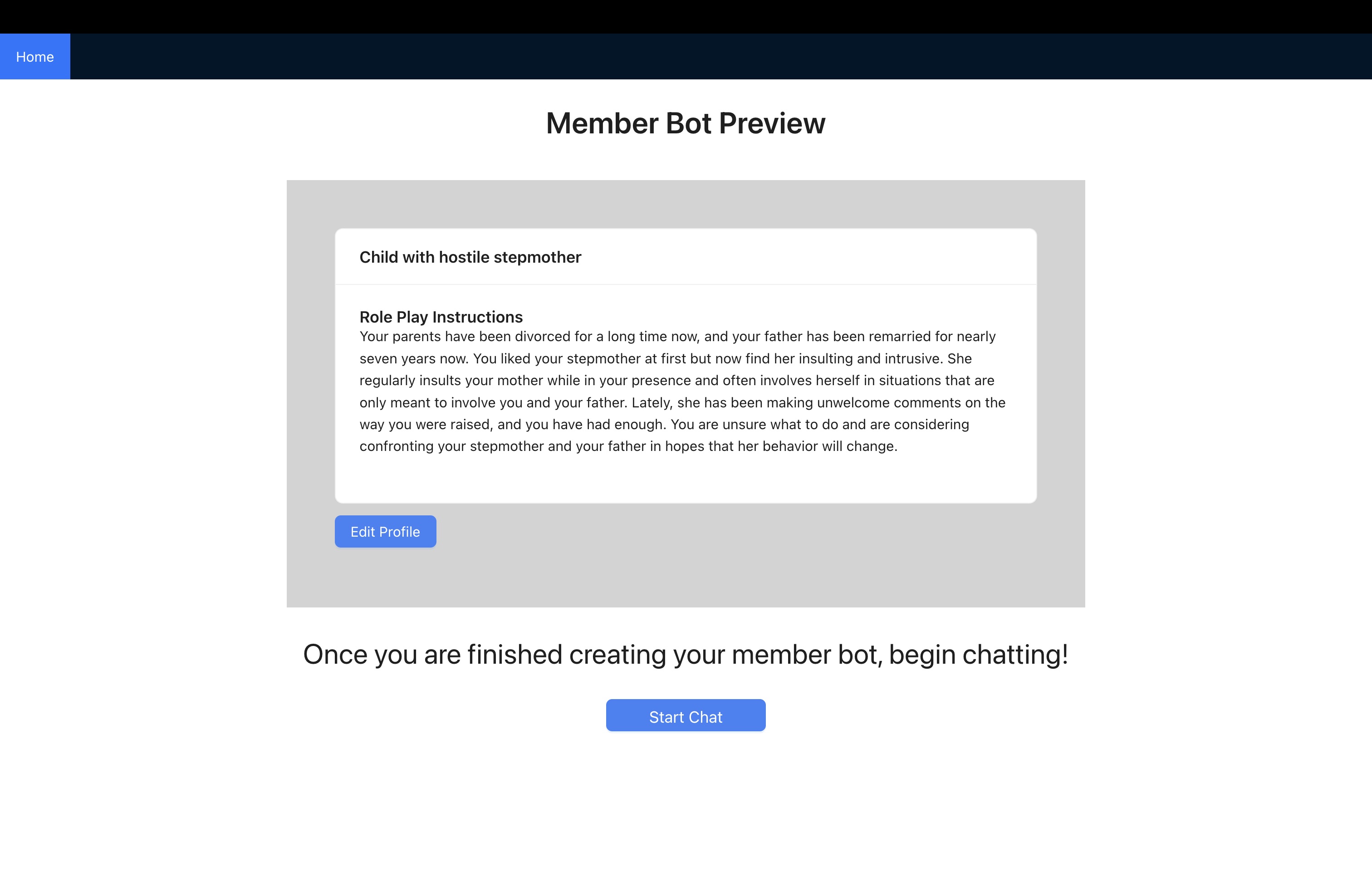}
    \caption{AI patient preview}
    \label{fig:screen4}
\end{figure*}

\begin{figure*}[ht]
    \centering
    \includegraphics[width=\textwidth]{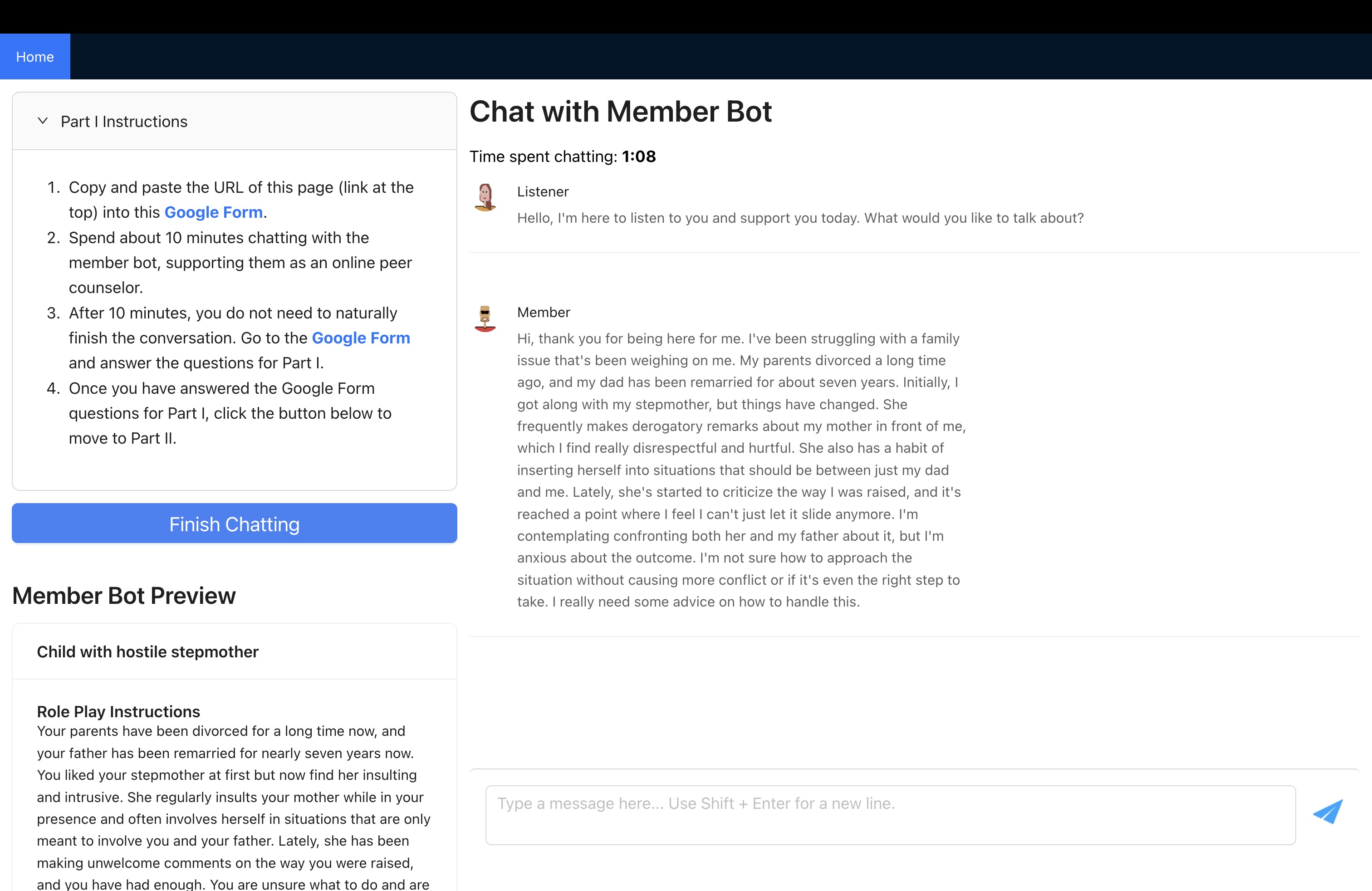}
    \caption{Part I chat with \textit{Scenario-Only} AI patient}
    \label{fig:screen5}
\end{figure*}

\begin{figure*}[ht]
    \centering
    \includegraphics[width=\textwidth]{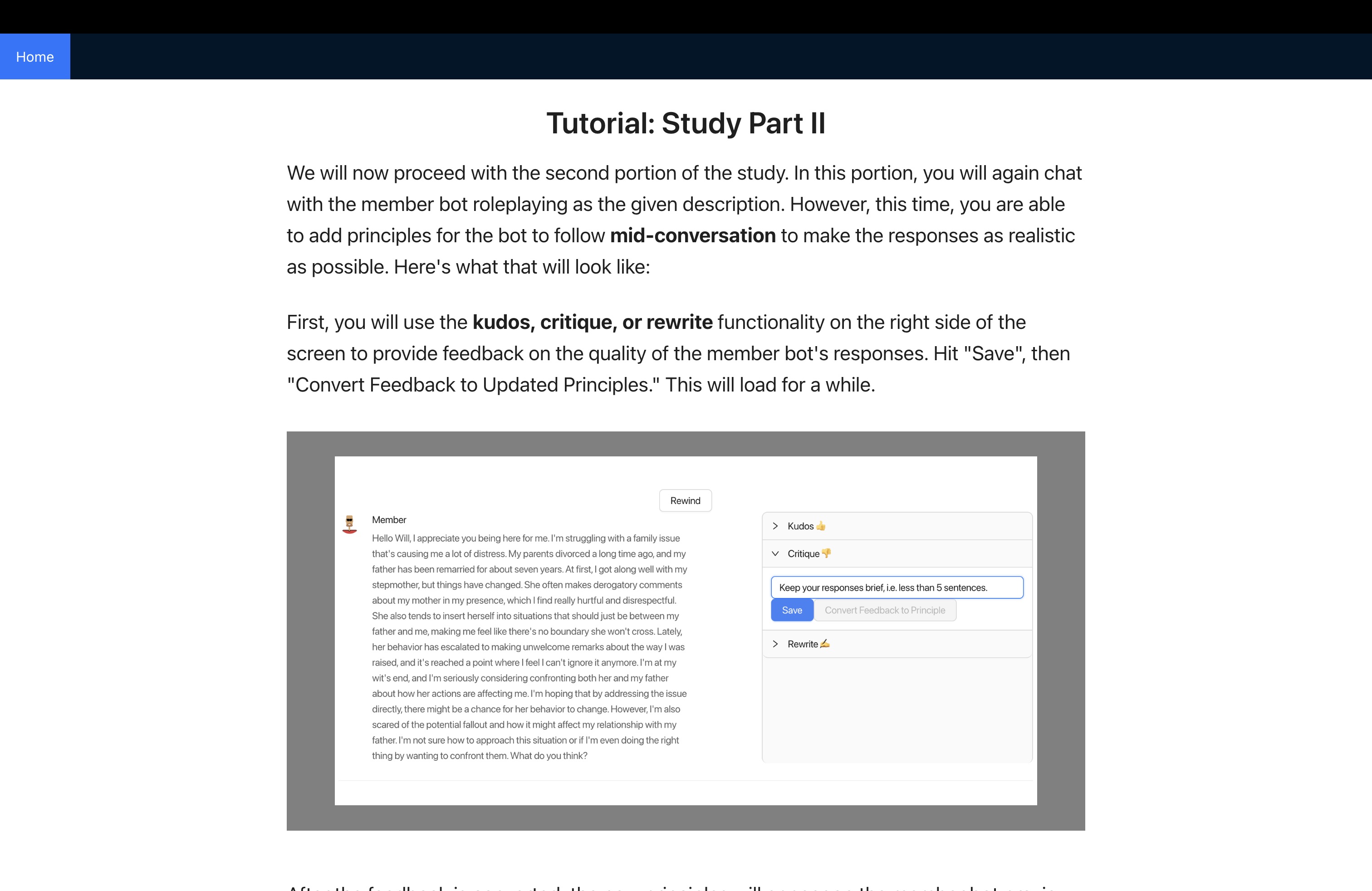}
    \caption{Part II instructions}
    \label{fig:screen6}
\end{figure*}

\begin{figure*}[ht]
    \centering
    \includegraphics[width=\textwidth]{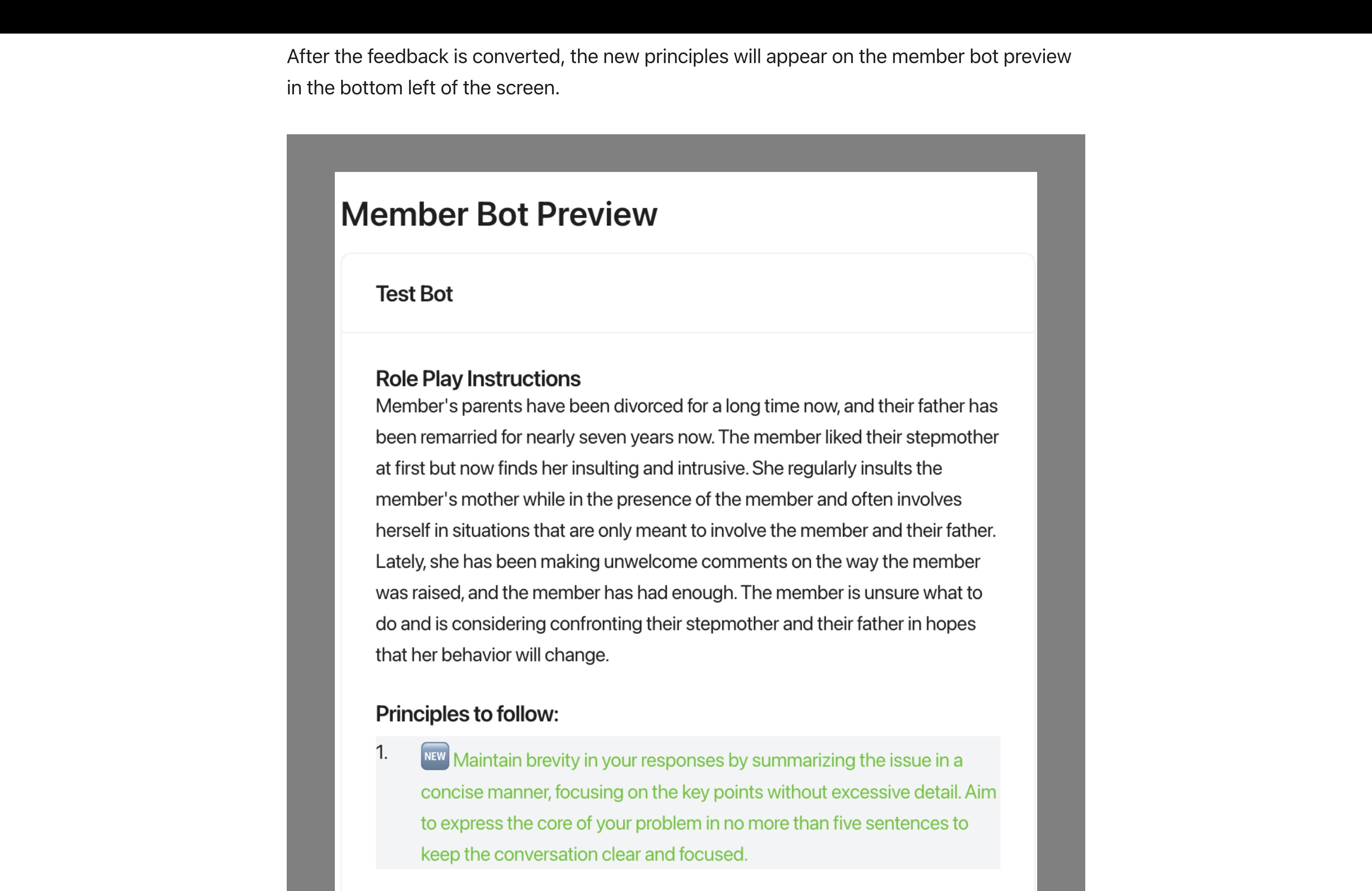}
    \caption{Part II instructions (continued)}
    \label{fig:screen7}
\end{figure*}

\begin{figure*}[ht]
    \centering
    \includegraphics[width=\textwidth]{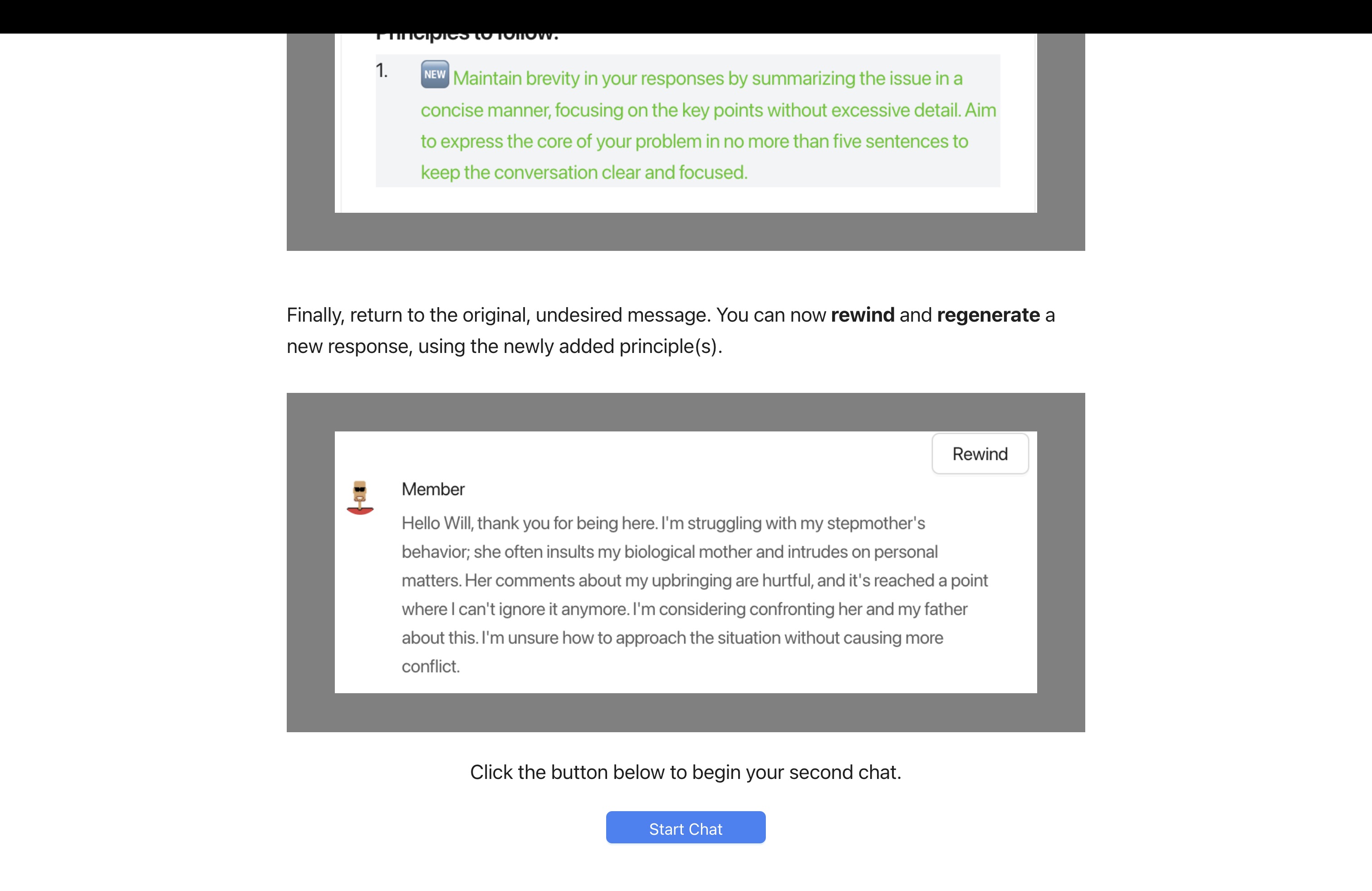}
    \caption{Part II instructions (continued)}
    \label{fig:screen8}
\end{figure*}

\begin{figure*}[ht]
    \centering
    \includegraphics[width=\textwidth]{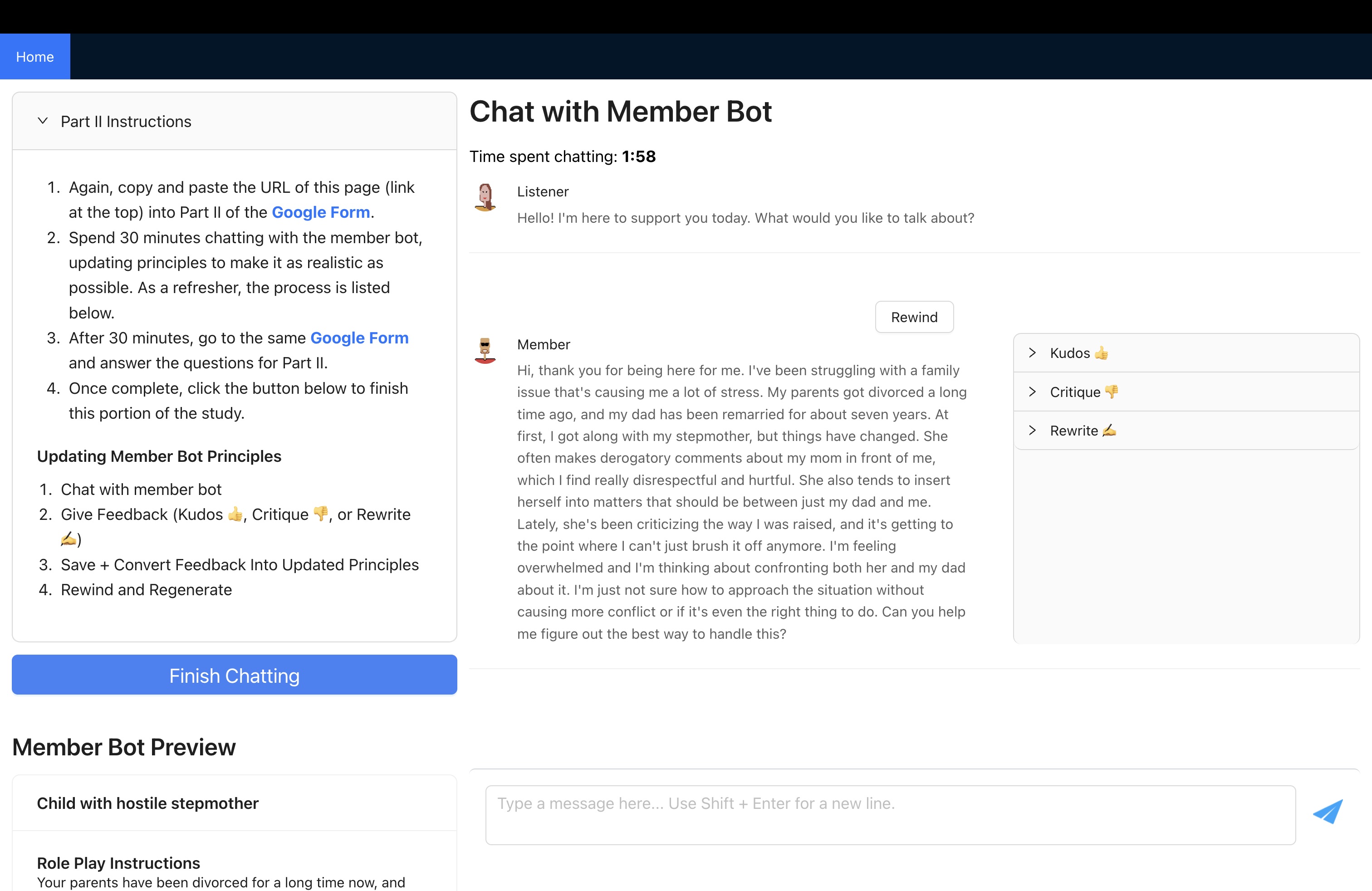}
    \caption{Part II chat with \textit{Scenario+Expert-Principles} AI patient}
    \label{fig:screen10}
\end{figure*}

\begin{figure*}[ht]
    \centering
    \includegraphics[width=\textwidth]{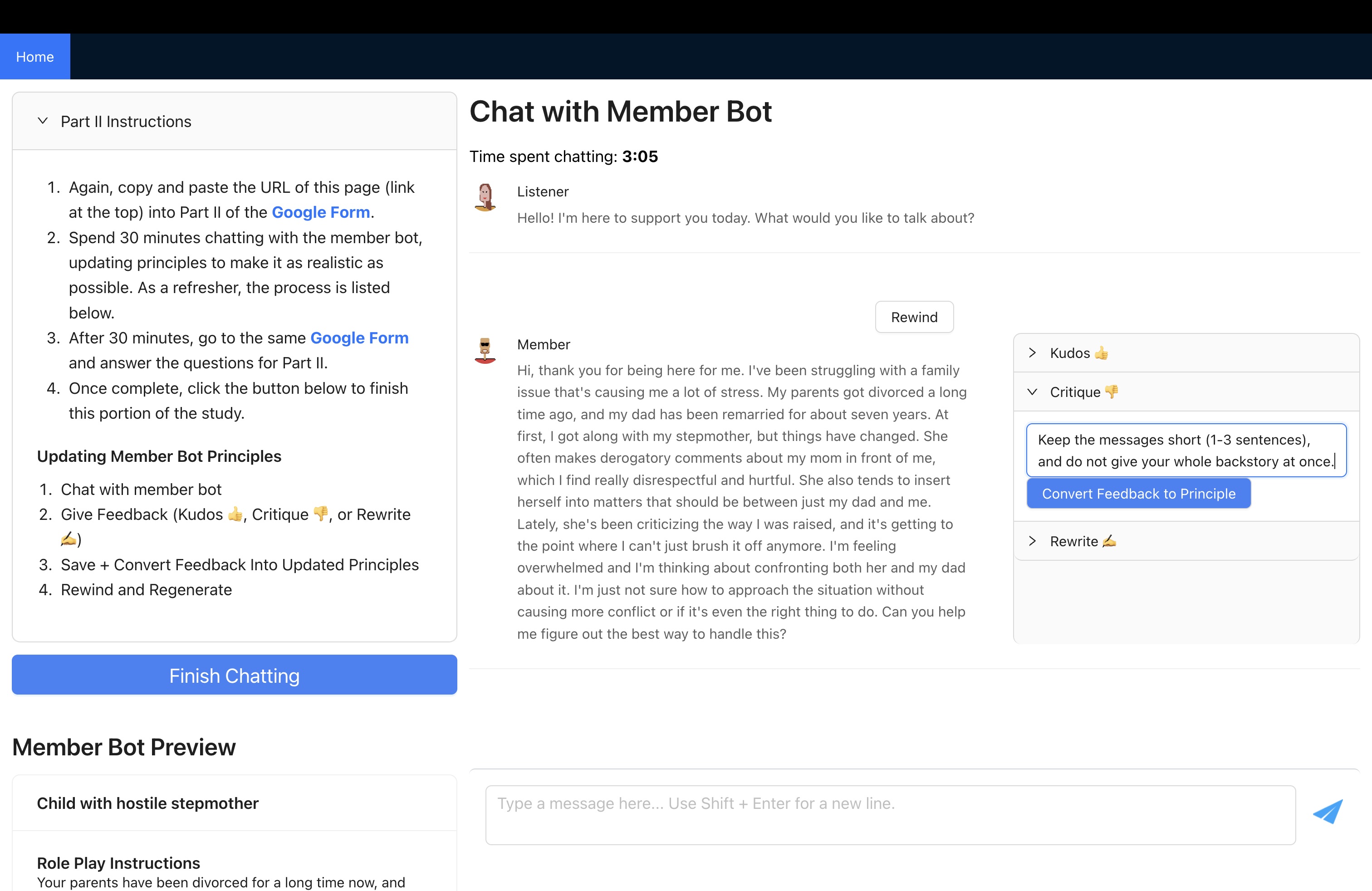}
    \caption{Using kudos/critique/rewrite to give feedback}
    \label{fig:screen11}
\end{figure*}

\begin{figure*}[ht]
    \centering
    \includegraphics[width=\textwidth]{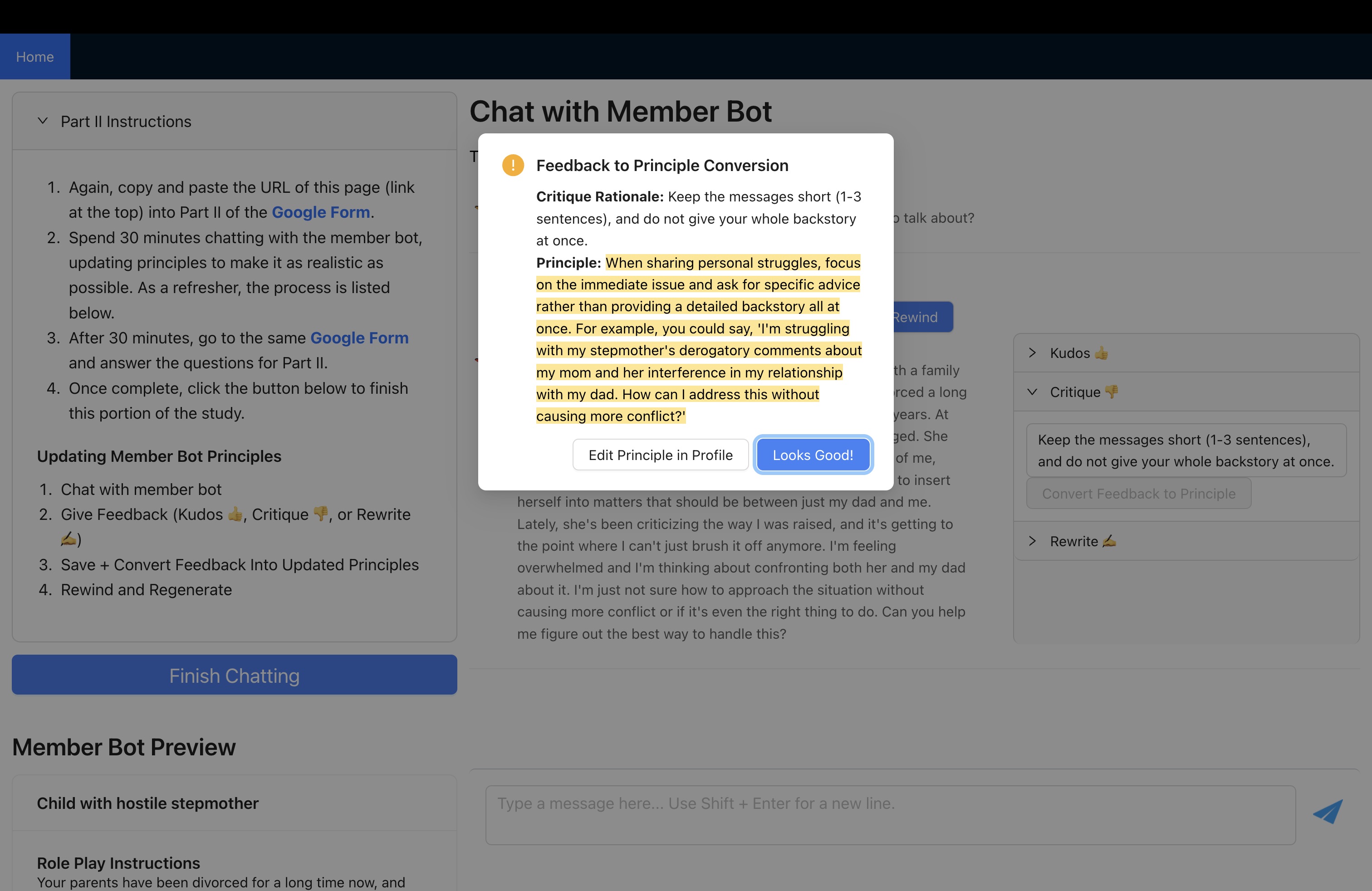}
    \caption{Feedback converted into principle}
    \label{fig:screen12}
\end{figure*}

\begin{figure*}[ht]
    \centering
    \includegraphics[width=\textwidth]{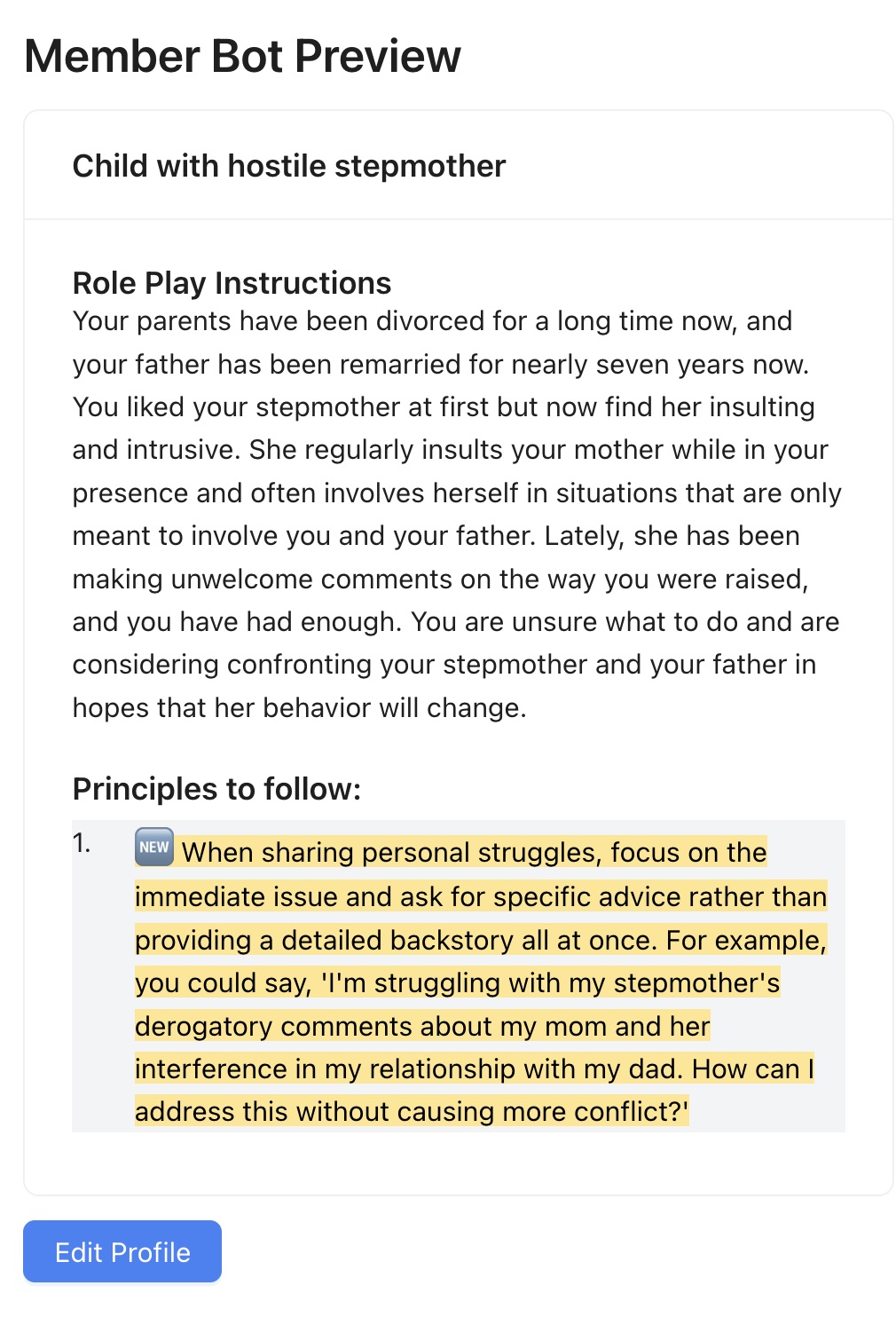}
    \caption{New principle incorporated into AI patient}
    \label{fig:screen13}
\end{figure*}

\begin{figure*}[ht]
    \centering
    \includegraphics[width=\textwidth]{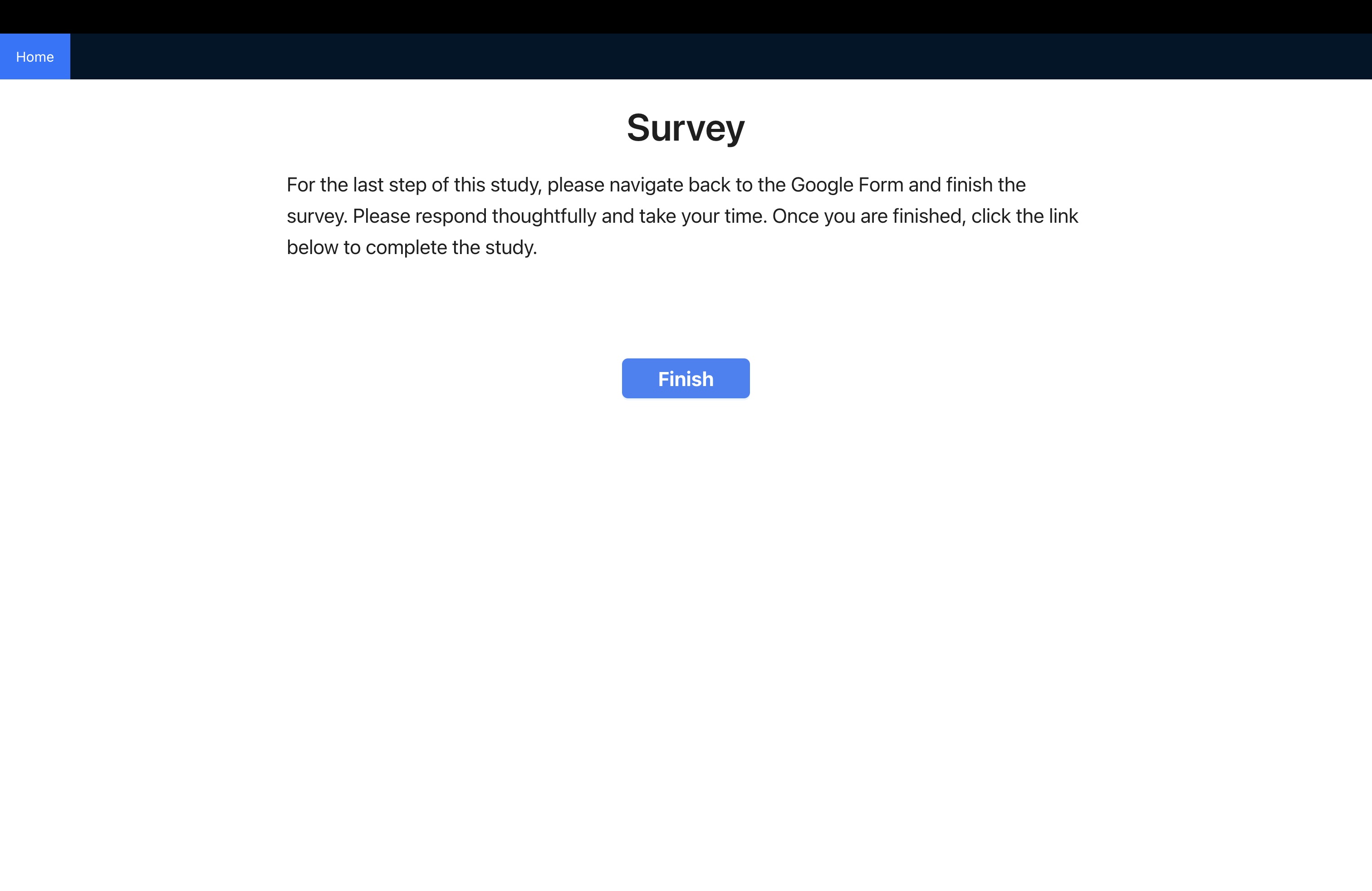}
    \caption{Finish and navigate to survey}
    \label{fig:screen14}
\end{figure*}

In this section, we describe the creator study flow that counselors followed during the 60-90 minute session. The reader can also refer to screenshots of our application that illustrates the different steps of this flow in Figures \ref{fig:screen1} to \ref{fig:screen14}.

Our study was designed to evaluate the impact of allowing counseling experts to add principles to Roleplay-doh on its perceived authenticity. We create a primarily self-guided study flow with accompaniment from the first author to clarify any points of confusion during the session.

To begin, participants first were introduced to the concept of AI patients used for training counseling skills in a simulated conversation. They were then instructed to write a challenging scenario that would serve as the scenario for the AI patients. 

The experimental procedure involved two main chat sessions. In Part I, participants engaged in a 10-minute conversation with the \textit{Scenario-Only} AI patient. Then, in Part II, participants interacted with the \textit{Scenario+Expert-Principles} AI patient for 30 minutes, keeping the same scenario from Part I and adding principles as the conversation progressed. After each of the two chat sessions, participants were asked to navigate to a form to evaluate the AI patients. 

\section{Creator Study Measures}
\label{appendix:creatorstudy-measure}

The following questions (Table \ref{tab:measures-roleplay} and \ref{tab:measures-tooluse}) are taken from the creator study questionnaire used to evaluate AI patients and the counselors' experience of using Roleplay-doh.
All items were rated on a 7-point Likert scale (1=Strongly disagree, 7=Strongly agree, except where noted below).
Table~\ref{tab:measures-roleplay} details the questions for evaluating the AI patient's roleplay, while Table~\ref{tab:measures-tooluse} details the questions about the experience using the tool to define principles.
Note that in the questions, we referred to the AI patients as ``Member Bots''. This terminology was used to match that of the online counseling platform 7 Cups, which refers to help seekers as ``Members'' within the support community.

\begin{table}[!h]
    \centering
    \resizebox{0.48\textwidth}{!}{
    \begin{tabular}{|c|p{0.6\columnwidth}|} \hline 
         Authenticity&The Member Bot in Part I/II played the role authentically.\\ \hline 
         Role Consistency& The Member Bot in Part I/II stayed in their role the whole time.\\ \hline 
         Resemblance to Case&How closely do you feel the conversation behaviors of the Member Bot in Part I/II resemble those of the specific past case you recall?\\ \hline 
         Challenging Aspects& Interacting with the Member Bot in Part I/II closely mirrored the challenging aspects I had experienced in the past case.\\ \hline
         Role readiness&The Member Bot in Part I/II is ready to be used as a simulated partner for training.\\ \hline 
         Recommend to novices&I would recommend the Member Bot from Part I/II to novice listeners/counselors to practice with.\\ \hline 
    \end{tabular}
    }
    \caption{Six measures used by creators to evaluate the two AI patients they created. Several measures were rephrased from prior work on evaluating Standardized Patients, or trained human actors, on case roleplay ability~\cite{himmelbauer2018standardized}.}
    \label{tab:measures-roleplay}
\end{table}
 
\begin{table}[!h]
    \centering
    \resizebox{0.48\textwidth}{!}{
    \begin{tabular}{|c|p{0.6\columnwidth}|} \hline 
         Effectively Guide& With the tool, I feel like I was able to write rules that can effectively guide the Member bot to recreate my past case.\\ \hline 
         Ease& With the tool, I felt like it was easy to convert my thoughts and feedback on the Member bot’s behavior into rules for the bot to follow.\\ \hline 
         Efficiency& With the tool, I felt like I could quickly and efficiently write rules for the bot.\\ \hline 
         Mental Demand& With the tool, I had to work very hard (mentally) to think of and write rules.\\ \hline

    \end{tabular}
    }
    \caption{Four measures as part of the tool usage section of the questionnaire taken from ~\cite{petridis2023constitutionmaker}}
    \label{tab:measures-tooluse}
\end{table}

\section{Worst-Case Analysis of Tool Experience} \label{sec:worst-case-tool-experience}
In a worst-case analysis of creators' tool experience, we uncovered cases where the human-LLM collaboration could be improved. Some counselors remarked that \textit{"having to think of and write rules was a challenge"} (P9) and that it \textit{"takes time to be specific"} when writing feedback (P7). Sometimes, even after giving feedback to the AI Patient, counselors like P19 observed that the patient \textit{"didn't always follow it"}, resulting in a non-progressive feedback loop, where \textit{"AI would generate [principles]... that were a little too similar to [feedback] I already gave, so that I was giving the AI the same feedback every time since it wasn't changing how it responded."}  While the principle-elicitation tools were designed to convert new feedback into a new principle, they operated ineffectively when follow-up feedback was given that was related to or a modification of previous feedback.  

As another issue, P23 noted the challenge in defining principles that generalize across specific contexts: \textit{"It was also hard to think about how to frame the feedback in an overarching way, rather than as direct feedback... directed as a specific part of the response"} (P24). While the principle-elicitation features aimed to help them convert specific feedback into generalized principles, imprecision in the feedback-to-principle conversion required counselors to edit the generalized-form of a principle in a way that was hard for them to articulate. 

These obstacles in tool experience could inspire future directions for improvement. First, to overcome issues in formulating rules, more support could be given to help those still unfamiliar with giving free-form feedback, such as through templates of feedback or principles that had high-success rates for past users. Second, to more seamlessly integrate follow-up feedback that is a clarification of previous feedback or principles, additional modules could help make sense of multiple pieces of feedback for the same response, and adopt LLM-assisted pipelines for user-driven criteria design~\cite{kim2024evallm} to support the merging of overlapping principles. Third, to overcome the abstraction gap between specific and abstract principles, more explicit representations that help to switch between specific and general feedback can be used.

\section{Third Party Study - Detailed Study Methods and Results}

\subsection{Third-party measures} \label{appendix-sec:thirdparty-detailed-measures}

Table~\ref{tab:measures-third-party} detail the six measures that third-party counselors answered for both AI patients.  Member Bot A and B refer to the AI patient whose transcript they read first and second, respectively.  Our analysis comparing \textit{Scenario-Only} and \textit{Scenario+ExpertPrinciples} accounts for this randomized the order of which AI patient they were shown.  

\begin{table}[!h]
    \centering
    \resizebox{0.48\textwidth}{!}{
    \begin{tabular}{|c|p{0.6\columnwidth}|} \hline 
         Authenticity&Member Bot A/B played the role authentically.\\ \hline 
         Role Consistency&Member Bot A/B stayed in their role the whole time.\\ \hline 
         Resemblance&Member Bot A's/B's behaviors closely mimicked the behaviors that typical clients/help-seekers exhibit.\\ \hline 
         Challenged Counselor& Member Bot A's/B's behaviors made it hard for the listener/counselor to give support.\\ \hline
         Role readiness&Member Bot A/B is ready to be used as a simulated partner for training.\\ \hline 
         Recommend to novices&I would recommend Member Bot A to novice listeners/counselors to practice with.\\ \hline 
    \end{tabular}
    }
    \caption{Six measures used by third-party counselors to judge the AI patients from an unbiased, external perspective.  Although the six dimensions largely overlap with those used in the creator study, the wording needed to be rephrased for the third-party perspective.}
    \label{tab:measures-third-party}
\end{table}

\subsection{Statistical Model and Power Analysis} \label{appendix-sec:third-party-power-analysis}

Via a power-analysis, we decided to recruit 5 counselors to act as external judges for 25-pairs of AI patients made in the creator study. In this section, we detail the procedures and results of this power-analysis. 

Generally, a power-analysis allows an experimenter to determine how many data-points are needed to detect a statistical difference for a particular effect size. Several prerequisites to conducting the power-analysis for the third-party study included (1) choosing a statistical model to test our hypothesis; and (2) estimating model parameters such as the effect of the treatment condition, the addition of  \textit{Expert Principles}, on annotator's ratings.

When choosing a statistical model as a pre-requisite, we needed a model that could account for how different annotators would be providing ratings to the same AI patients created by each counselor. A traditional paired t-test was not appropriate because the independent samples assumption is violated due to different annotators giving ratings to the same AI patients. While another common practice is using the majority vote between annotators, our pilot data found that annotators did not always have high agreement. Therefore, since we wanted to account for the variability between annotators as well as between the ratings, we chose to use a linear mixed-effects model.  Using the \lstinline!lme4! package in R \cite{bates2015package}, this model is defined as \lstinline!Rating~Treatment+CreatorID+(1|AnnotatorID)!.  This model defines the treatment group (whether the AI patient has Expert Principles or not) as fixed effects, the creator ID's as fixed effects to account for the pair of AI patients made by each counselor, and the annotators as random effects. This approach can handle the non-independence of annotator ratings.

Prior to performing the power analysis, we needed to define the expected parameters of this linear mixed effect model. To define these expected parameters, we fit a model to early study data in which 2 annotations had been collected for each pair of AI patients created by 17 counselors. Specifically, we extracted the fixed effects, the random effects covariance matrix, and residual variances. 

\begin{figure}[t]
    \centering
    \includegraphics[width=\columnwidth]{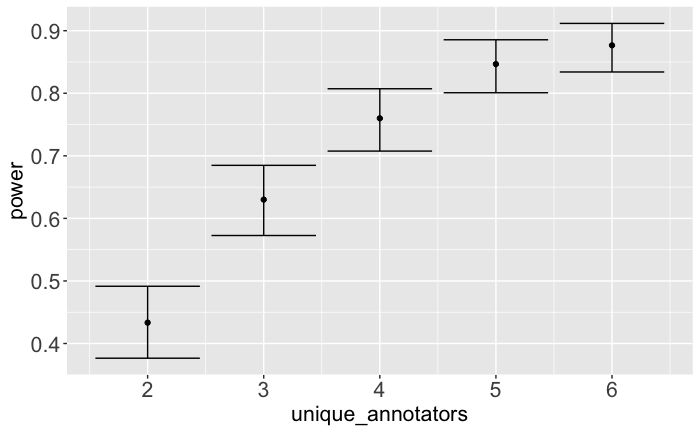}
    \caption{Based on our simulation-based power analysis across 300 trials for our linear, mixed-effect model, we conclude that 80\% power can be achieved with 5 third-party judges.}
    \label{fig:power-analysis}
\end{figure}

A simulation-based approach is the most feasible method for doing power-calculations for mixed-effect models. In this approach, an experimenter simulates data based on specified parameters (effect sizes, variance components, sample sizes) and analyzes the data repeatedly to estimate power empirically. We used the \lstinline!simr! package in R to conduct a simulation-based power-analysis~\cite{green2016simr}. In the power-analysis, we varied how many unique annotators from 2 - 6 to understand the frequency of trials which would detect a treatment effect of $0.52$ at significance-level $\alpha=0.05$. Our simulation-based power-analysis over 300 trials are shown in Figure~\ref{fig:power-analysis}. We concluded that we could achieve greater than 80\% power using 5 judges. 

\subsection{Why is the effect of Expert Principles smaller when rated by a third-party?} \label{appendix-sec:thirdparty-qual-analysis}

Here we further investigate how third-party annotators rated each of the 25-pairs of AI patients created in our study. In particular, we investigate why the effect of \textit{ExpertPrinciples} is lower than what was measured in the creator study from a first-person perspective. 

One reason for this smaller effect is the lower agreement between the 5 third-party counselor. Specifically, we used the Krippendorf $\alpha$ metric to compute the agreement for the comparisons of the AI patients with and without expert principles (Table~\ref{tab:thirdparty_ratingdifference_alpha}). Across the 6 measures, we find that Krippendorf's $\alpha$ is between 0.046-0.232 for which patient they prefer, indicating between low to moderate agreement. 

Third-party raters also provided rationales which helped us better understand their thought process. We filtered cases in which there is a disagreement between third-party counselors on which AI patient is better, and investigated these rationales. \textbf{We find that counselors note similar behaviors in the AI patient, meaning they agree on their observations}. For example, for the AI patient created by P3, both third-party annotators observed that the AI patient based on the \textit{Scenario-only} resolved their problems too quickly, whereas the AI Patient with \textit{ExpertPrinciples} added allows the \textit{"listener to ask questions and explore with the client"}. However, the third-party annotator that prefers \textit{Scenario-only} stated that the \textit{Scenario+ExpertPrinciples} patient sounded too \textit{formulaic and robotic, whereas the other is more expressive and realistic}. Looking further into what the creator said about this AI patient, they mentioned that the \textit{Scenario+ExpertPrinciples} patient \textit{talks like an actual person would... there's a good balance of going into just enough detail on noting experiences, describing struggles, while maintaining the brevity.} What this case illustrates is that \textbf{different counselors can disagree on what principles are the most relevant for an authentic roleplay.}

\begin{table*}[t]
    \small
    \centering
    \begin{tabular}{|c|l|l|} \hline  
         \textbf{Metric}&  \textbf{$\alpha$ for Rating Difference}&\textbf{$\alpha$ for Preference}\\ \hline  
         Authenticity&  0.043 &0.089\\ \hline  
 Stayed in Role&0.023 &0.046\\ \hline 
         Resemblance&   0.076 &0.148\\ \hline  
         Mirrors Challenges&   0.041 &0.085\\ \hline  
         Ready&   0.075 &0.209\\ \hline  
         Recommend&   0.082 &0.232\\ \hline 
    \end{tabular}
    \caption{Krippendorff's $\alpha$ for the comparisons made between \textit{Scenario+ExpertPrinciples} and the \textit{Scenario-Only} AI patients, as judged by third-party counselors. We compute $\alpha$ for both the rating difference, and the preference (i.e. signed rating difference) between the two AI patients.}
    \label{tab:thirdparty_ratingdifference_alpha}
\end{table*}

\subsection{Creator Study Conversation Lengths}

In Table~\ref{tab:descriptivestats-convos}, we show descriptive statistics of the conversations collected during the user studies between creators and AI patients.
\begin{table*}[t]
\small
\centering
\begin{tabular}{|c|c|c|c|c|}
\hline
{\textbf{Participant}} & 
{\textbf{\# Utterances (Part 1)}} & 
{\textbf{\# Utterances (Part 2)}} & 
{\textbf{Mean Output Length (Part 1)}} & 
{\textbf{Mean Output Length (Part 2)}} \\
\hline
1  &  8 &  6 & 114.75  & 169.00  \\ \hline
2  & 18 & 19 & 235.89  & 278.40  \\ \hline
3  & 10 & 18 & 255.45  & 112.56  \\ \hline
4  & 14 & 14 & 161.86  & 62.14   \\ \hline
5  & 12 &  6 & 201.00  & 149.33  \\ \hline
6  & 10 &  9 & 133.80  & 46.00   \\ \hline
7  &  8 & 10 & 162.00  & 123.40  \\ \hline
8  & 12 &  8 & 145.33  & 113.50  \\ \hline
9  &  6 & 12 & 269.67  & 103.33  \\ \hline
10 & 10 & 12 & 168.20  & 158.33  \\ \hline
11 &  8 & 10 & 110.00  &  41.40  \\ \hline
12 & 12 &  8 & 131.50  &  70.75  \\ \hline
13 & 12 & 10 & 164.50  &  65.60  \\ \hline
14 & 20 & 14 &  34.00  &  25.86  \\ \hline
15 & 12 & 11 & 117.17  &  75.00  \\ \hline
16 & 14 & 18 & 162.14  &  69.80  \\ \hline
17 & 12 & 18 & 259.83  &  91.55  \\ \hline
18  & 16 & 26 & 240.25 & 79.92 \\ \hline
19  & 14 & 16 & 254.71 & 243.88\\ \hline
20  & 12 & 14 & 144.00 & 106.00\\ \hline
21   & 20 & 21 & 125.00 & 159.81\\ \hline
22  & 18 & 12 & 120.44 & 245.00 \\ \hline
23  & 12 & 14 & 231.67 & 147.42\\ \hline
24  & 14 & 22 & 184.71 & 142.45\\ \hline
25  & 22 & 12 & 304.00 & 130.00\\ \hline
\textbf{Mean} & \textbf{13.04} & \textbf{13.64} & \textbf{177.29} & \textbf{120.43} \\
\hline
\end{tabular}
\caption{Descriptive statistics per conversation with the two versions of the AI Patient in Part 1 (\textit{Scenario-Only}) and Part 2 (\textit{Scenario+ExpertPrinciples}). Output length is measured in number of tokens.}
\label{tab:descriptivestats-convos}
\end{table*}

\section{Detailed Results for principle-adherence pipeline Evaluation}
\label{sec:detres}

\begin{table*}[h!]
\small
\centering
\begin{tabular}{|c|c|c|c|c|}
\hline
Method                    & Metric 1 & Metric 2 & Metric 3 & Overall Ranking \\ \hline
Full                      & 0.257  & 0.484    & 0.208       & 0.444           \\ \hline
Naive                     & 0.543   & 0.538    & 0.644       & 0.786           \\ \hline
No Principle Rewrites     & 0.278   & 0.302    & 0.411       & 0.528           \\ \hline
No Autogenerated Criteria & 0.387   & 0.608    & 0.492       & 0.592           \\ \hline
No Critique & -   & 0.562    & -       & -           \\ \hline
\end{tabular}
\caption{Krippendorff's $\alpha$ for error testcases across metrics and methods.}
\label{tab:alpha2}
\end{table*}

\begin{table*}[h!]
\small
\centering
\begin{tabular}{|c|c|c|c|c|}
\hline
Method                    & Metric 1 & Metric 2 & Metric 3 & Overall Ranking \\ \hline
Full                      & 0.229    & 1.0   & 0.226       & 0.440           \\ \hline
Naive                     & 0.362  & 1.0     & 0.607       & 0.747           \\ \hline
No Principle Rewrites     & 0.202  & 1.0     & 0.130       & 0.311           \\ \hline
No Autogenerated Criteria & 0.169   & 1.0    & 0.174       & 0.498           \\ \hline
No Critique & -   & 1.0    & -       & -           \\ \hline
\end{tabular}
\caption{Krippendorff's $\alpha$ for random testcases across metrics and methods.}
\label{tab:alpha1}
\end{table*}

We first provide Krippendorff's $\alpha$ numbers for inter-annotator agreement in Table \ref{tab:alpha1} and \ref{tab:alpha2} for both the random and error testcases. The random testcases are 50 randomly picked conversation turns from the user study logs, and the experiment detailed in Section \ref{sec:evalpap} is carried out on them. We find that agreement scores lie in the 0.2-0.6 range, indicating fair agreement between annotators.

\begin{figure*} [ht]
    \centering
    \includegraphics[width=\textwidth]{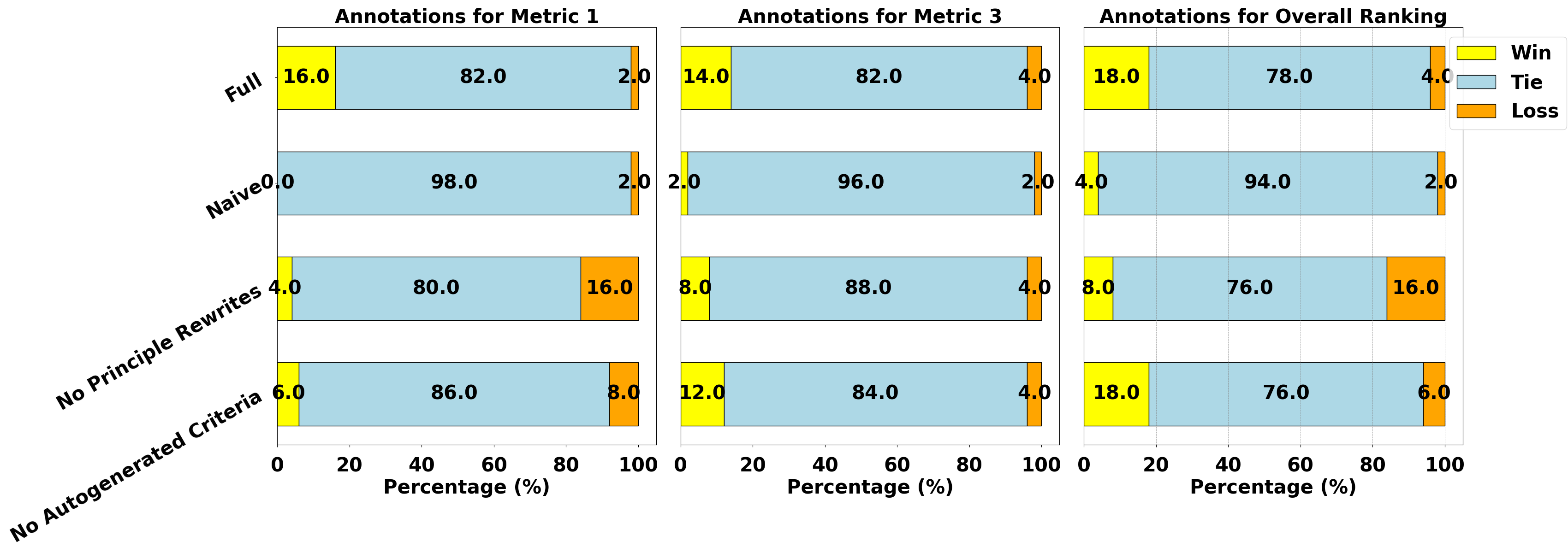}
    \caption{Win/Tie/Loss for the Random Test Cases along \textbf{M1}, \textbf{M3}, and \textbf{Overall}.}
    \label{fig:wtl-error2}
\end{figure*}

Next, we provide results for our evaluation study on the random testcases in Figure \ref{fig:wtl-error2}. We observe a substantial increase in tie rate across modules and metrics \textbf{M1} and \textbf{M3} as well as the overall ranking. This is expected because a relatively small proportion of responses from [$\textbf{No Critique}$] contain errors that should be corrected by the principle-adherence pipeline. In these cases, we expect the no rewrites, or the rewritten response being of similar quality to the original response. However, we still find that our [\texttt{Full}]
method performs better than [\texttt{No Critique}] on
\textbf{M1} (W: 15\%; L 2\%) and on M3 (W: 14\%; L 4\%), where it has the highest win/loss rates compared to all ablations. This hold true for overall ranking as well (W: 18\%; L 4\%). This highlights that our [\texttt{Full}] approach results in improved quality of responses even when the proportion of erros is relatively low. For \textbf{M2}, all annotators report no awkward responses for all methods. 

\section{Annotation Interface for principle-adherence pipeline Evaluation}
\label{sec:annotint}

\begin{figure*}
    \centering
    \includegraphics[width=\textwidth]{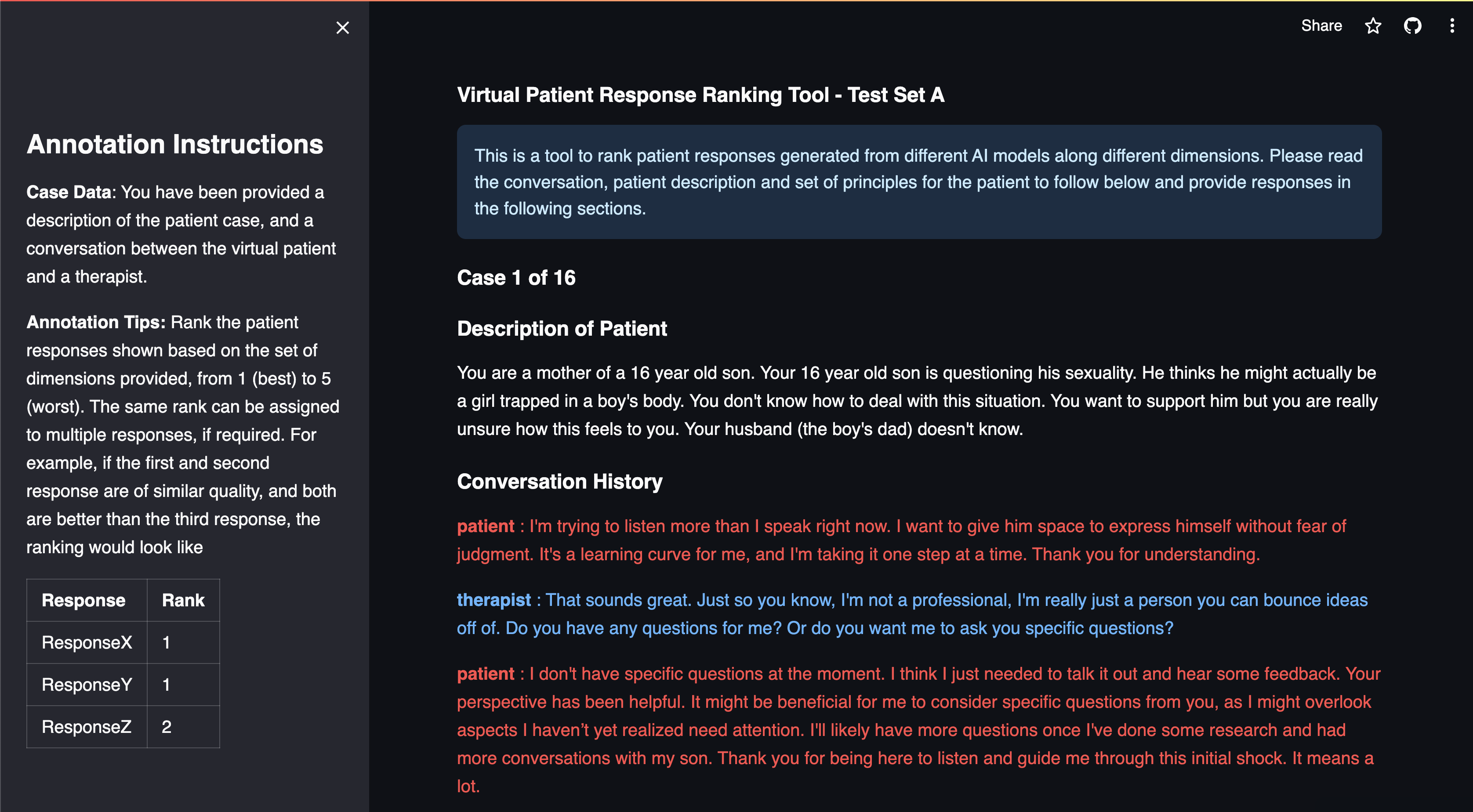}
    \caption{Principle Adherence Annotation Interface: Case Input with Patient Description and Conversation History}
    \label{fig:ranking-interface-caseinput}
\end{figure*}

\begin{figure*}
    \centering
    \includegraphics[width=\textwidth]{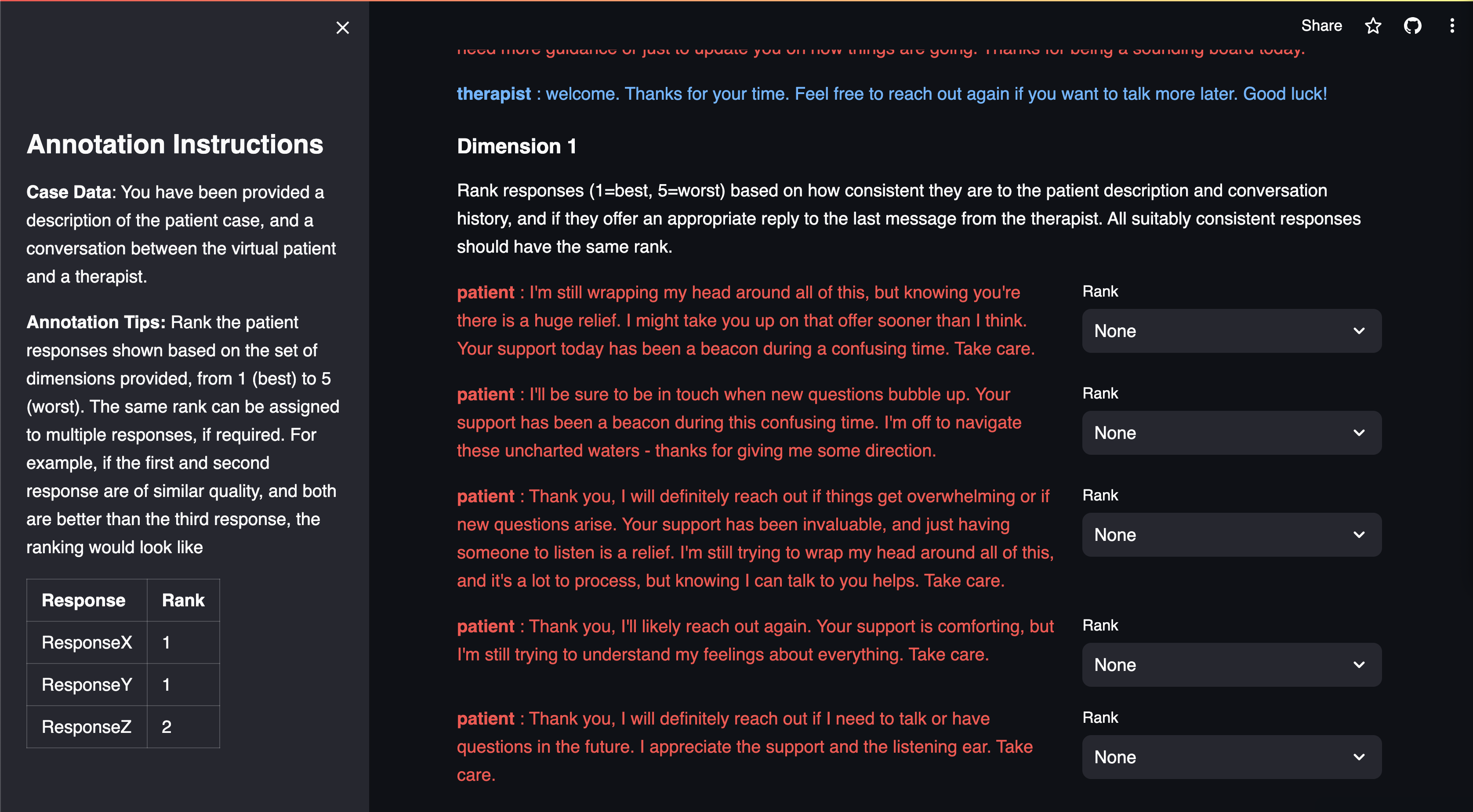}
    \caption{Principle Adherence Annotation Interface: Questions to get annotations for \textbf{M1}, or consistency in dialogue history.}
    \label{fig:ranking-interface-m1}
\end{figure*}

\begin{figure*}
    \centering
    \includegraphics[width=\textwidth]{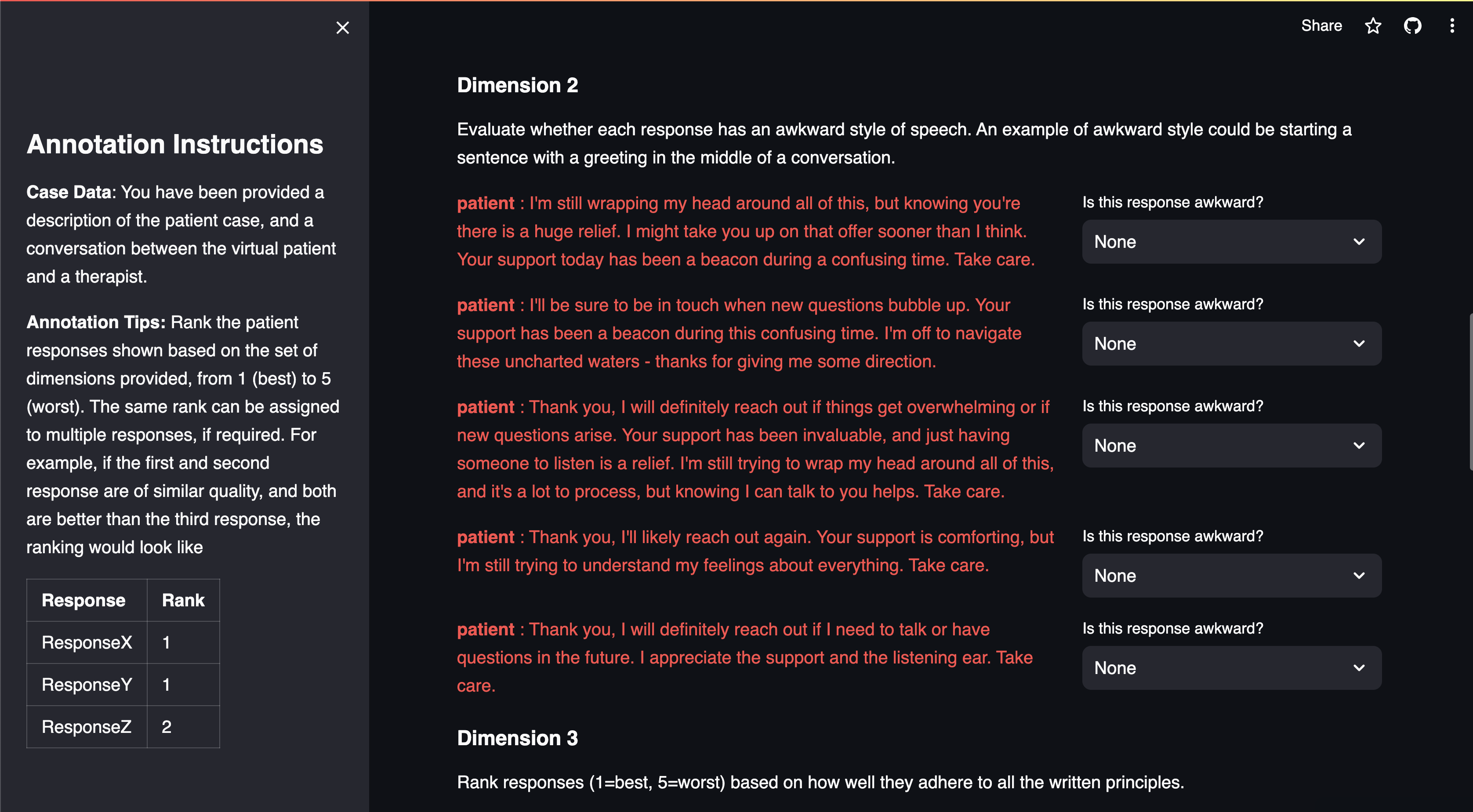}
    \caption{Principle Adherence Annotation Interface: Questions to get annotations for \textbf{M2}, or awkwardness in responses.}
    \label{fig:ranking-interface-m2}
\end{figure*}

\begin{figure*}
    \centering
    \includegraphics[width=\textwidth]{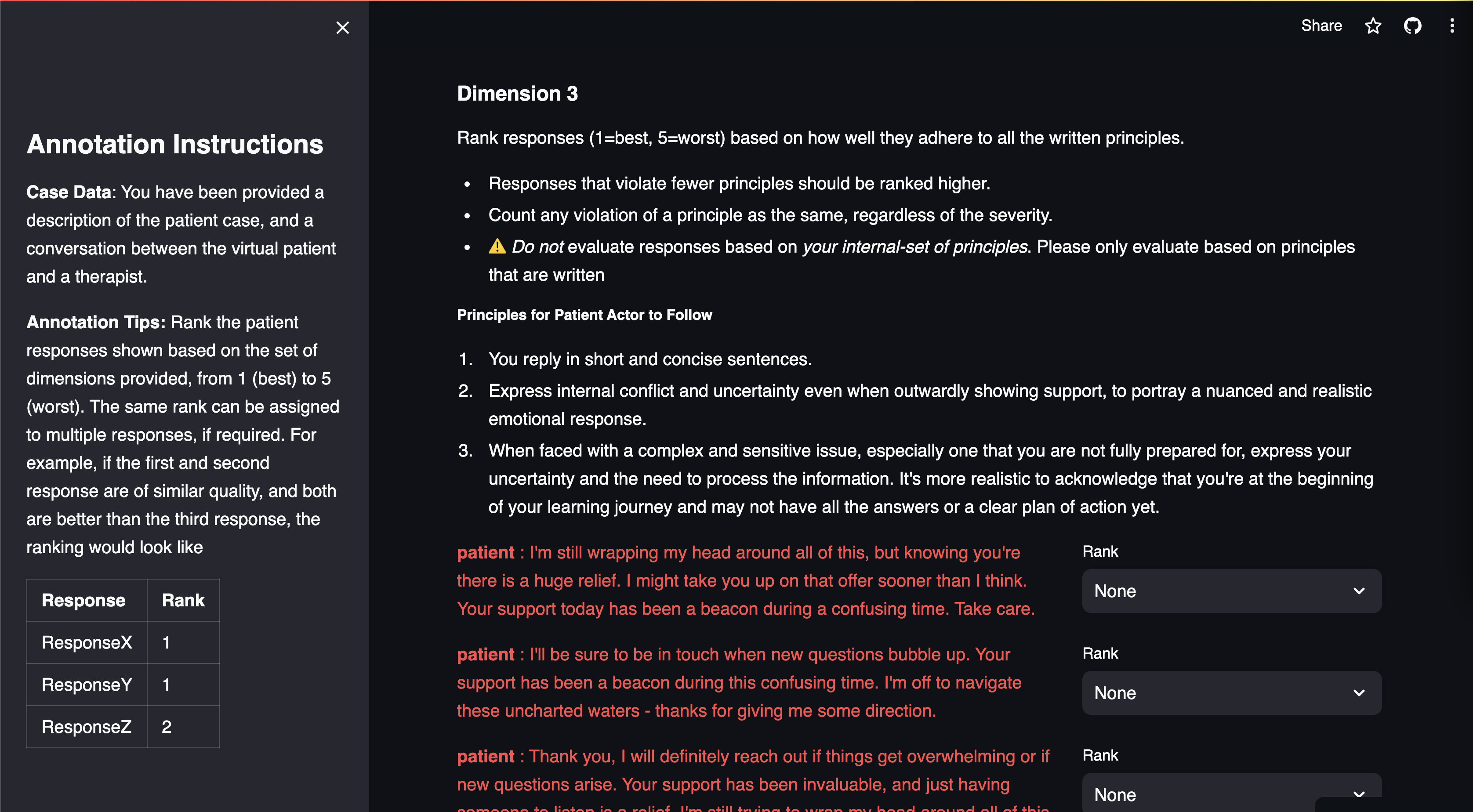}
    \caption{Principle Adherence Annotation Interface: Questions to get annotations for \textbf{M3}, or adherence to all written principles.}
    \label{fig:ranking-interface-m3}
\end{figure*}

\begin{figure*}
    \centering
    \includegraphics[width=\textwidth]{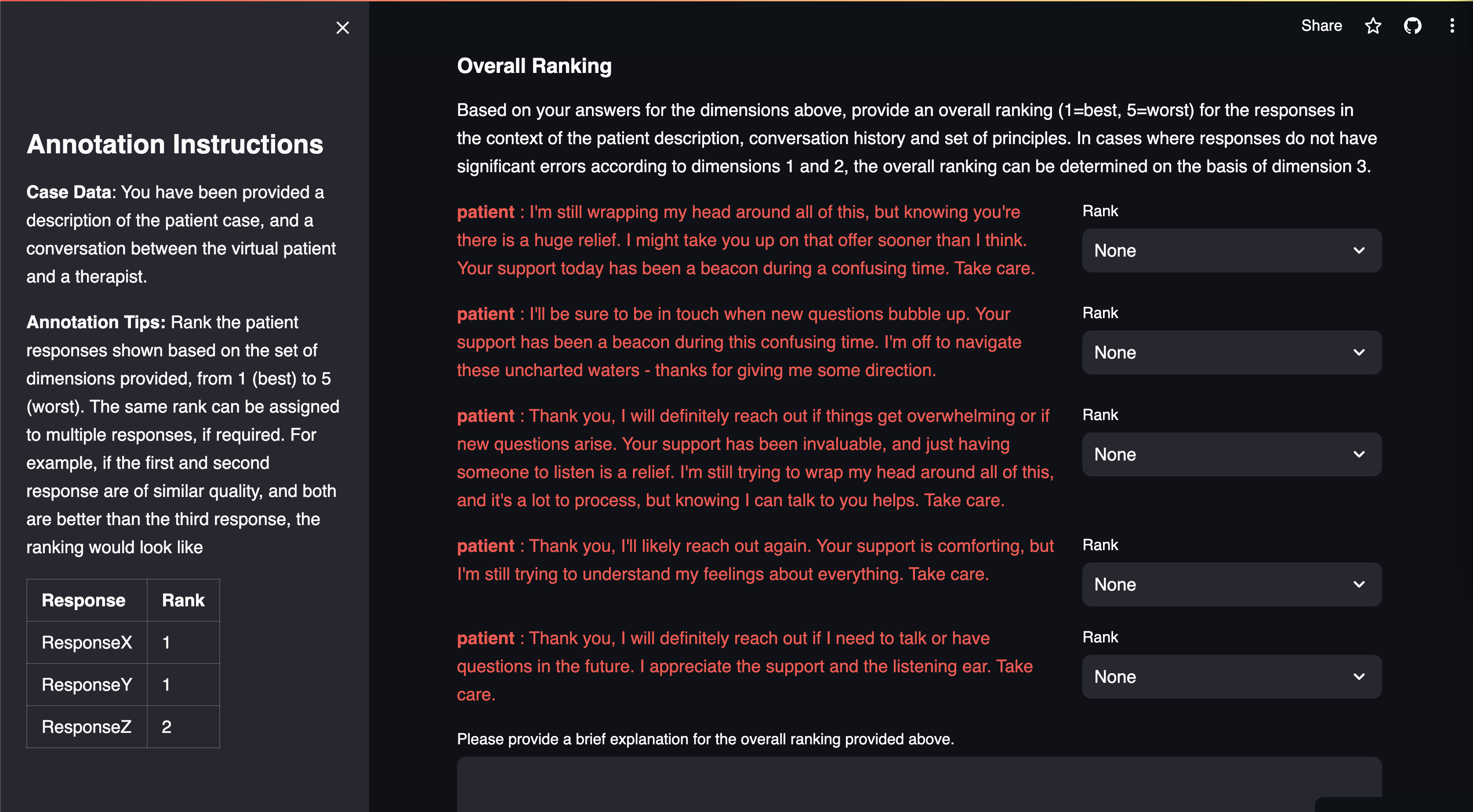}
    \caption{Principle Adherence Annotation Interface: Questions to get annotations for an \textbf{Overall} ranking, which also includes a free text field to capture a rationale.}
    \label{fig:ranking-interface-overall}
\end{figure*}

Figures \ref{fig:ranking-interface-caseinput}, \ref{fig:ranking-interface-m1}, \ref{fig:ranking-interface-m2}, \ref{fig:ranking-interface-m3} and \ref{fig:ranking-interface-overall} provides an overview of the annotation interface used in the principle-adherence evaluation study. In certain cases, multiple methods resulted in the same output for a testcase. These responses are deduplicated before presenting to the user. Ranks assigned to the duplicated response are then assigned to all models that resulted in the response. Notable, in 34/50 of the random testcases, all models resulted in the same response. These testcases were not annotated, and a rank of 1 was assigned to all models. These cases are also not considered while calculating Krippendorff's $\alpha$ in Appendix \ref{sec:detres}.

\end{document}